\documentclass[a4paper,fleqn]{cas-dc}

\usepackage{soul} 
\usepackage{color, xcolor}
\definecolor{light-gray}{gray}{0.90}

\usepackage[numbers]{natbib}

\usepackage{subfigure}
\usepackage{tabularx}
\usepackage{tikz}
\usepackage{pgfplots}
\usepackage[edges]{forest}
\usepackage{adjustbox}

\usepackage{makecell}

\usepackage{tabularx}

\newtheorem{definition}{Definition}[section]
\counterwithout{definition}{section}

\newtheorem{principle}{Principle}[section]
\counterwithout{principle}{section}

\usepackage[switch]{lineno}

\begin{document}

\shorttitle{Graph Data Augmentation with Contrastive Learning on Covariate Distribution Shift}
\shortauthors{F. Zeng \textit{et al.}}

\author[1]{Fanlong Zeng}
\ead{flzeng1@stu.jnu.edu.cn}
\address[1]{School of Intelligent Systems Science and Engineering, Jinan University, Zhuhai 519070, China}

\author[1]{Wensheng Gan}
\cortext[cor1]{Corresponding author}
\ead{wsgan001@gmail.com}
\cormark[1]

\title[mode = title]{Graph Data Augmentation with Contrastive Learning on Covariate Distribution Shift}

\begin{abstract}
  Covariate distribution shift occurs when certain structural features present in the test set are absent from the training set. It is a common type of out-of-distribution (OOD) problem, frequently encountered in real-world graph data with complex structures. Existing research has revealed that most out-of-the-box graph neural networks (GNNs) fail to account for covariate shifts. Furthermore, we observe that existing methods aimed at addressing covariate shifts often fail to fully leverage the rich information contained within the latent space. Motivated by the potential of the latent space, we introduce a new method called MPAIACL for \underline{M}ore \underline{P}owerful \underline{A}dversarial \underline{I}nvariant \underline{A}ugmentation using \underline{C}ontrastive \underline{L}earning. MPAIACL leverages contrastive learning to unlock the full potential of vector representations by harnessing their intrinsic information.  Through extensive experiments, MPAIACL demonstrates its robust generalization and effectiveness, as it performs well compared with other baselines across various public OOD datasets. The code is publicly available at \href{https://github.com/flzeng1/MPAIACL}{https://github.com/flzeng1/MPAIACL}.
\end{abstract}


\maketitle

\section{Introduction}

Graph classification \cite{ju2024towards, sui2024unleashing} is a fundamental task in real-world graph analysis, distinguished from other classification tasks by its reliance on node and edge representations to capture the complex structure and semantics of entire graphs \cite{ju2024towards}. Graph neural networks (GNNs) \cite{velivckovic2018graph} have recently emerged as the key framework for modeling graph-structured data. Currently, GNNs are typically designed under the assumption that the training and test sets are drawn from independent and identically distributed (I.I.D) data \cite{arjovsky2019invariant}. However, this assumption barely holds in real-world scenarios, due to the out-of-distribution (OOD) problem potentially existing during the test stage \cite{sui2024unleashing}. Additionally, the OOD problem induces a distribution shift between the training and test sets, leading to a significant degradation in model performance. When applied to datasets with distribution shifts, GNNs typically yield suboptimal overall graph classification performance \cite{gui2022good}. As a result, various approaches have been proposed to address this challenge, such as invariant learning \cite{sui2022causal}, architecture design \cite{yang2022graph}, and data augmentation \cite{han2022g}.

\begin{figure}[]
    \centering
    \vspace{0.6cm} 
    \includegraphics[scale=0.6]{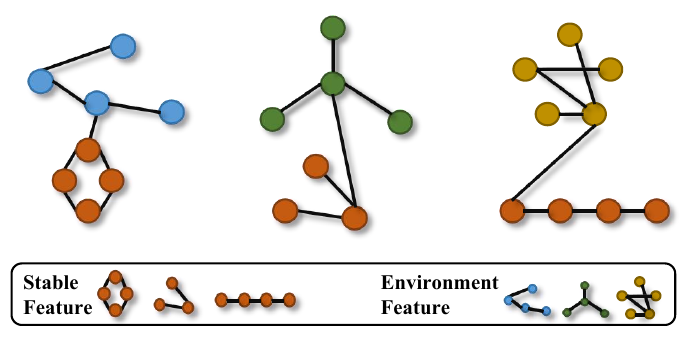}
    \caption{The stable features and environment features. Stable features capture the underlying patterns of the entire graph, providing a robust representation of the graph's intrinsic structure. Environmental features are subject to variation for the label.}
    \label{fig: concept of stable and env features}
\end{figure}

\begin{figure*}
    \label{fig: difference between covariate and correlation}
    \centering
    \subfigure[Correlate distribution shift. Correlation shift denotes that environment features establish the spurious relation with labels.]{
        \label{fig: correlate shift}
        \includegraphics[scale = 0.6]{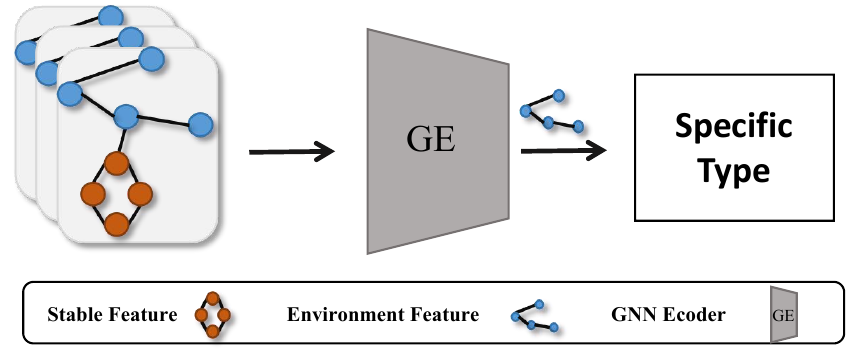}
    }
    \subfigure[Covariate distribution shift. Covariate shift denotes that environment features are very different between the training set and test set.]{
        \label{fig: covariate shift}
        \includegraphics[scale = 0.6]{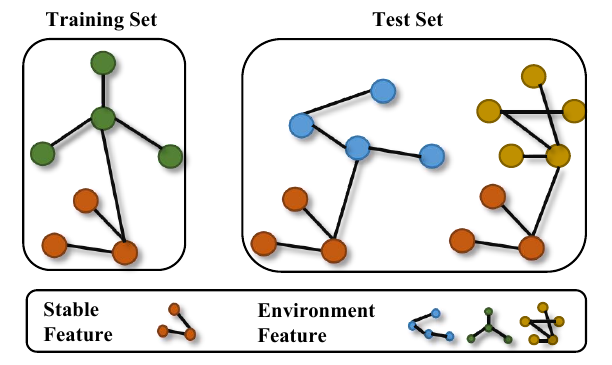}
    }
    \caption{Correlate distribution shift and covariate distribution shift.  }   
\end{figure*}

The distribution shift can be further categorized \cite{sui2024unleashing}. For a better understanding of these distribution shift problems, we introduce two important concepts, which are the stable feature and the environment feature.
According to previous studies \cite{liu2022graph}, we have the concepts below:
i) Stable features,  which capture the underlying patterns of the entire graph, providing a robust representation of the graph's intrinsic structure. Based on this, the relationship between stable features and labels can be considered invariant; ii) Environmental features, which are subject to variation for the label \cite{sui2024unleashing}. We can consider a toy example of a molecule. As illustrated in Figure \ref{fig: concept of stable and env features}, the stable feature is a component that can determine the property of a molecule, like functional groups. While the environment features are various and irrelevant to the property. Noticed that functional groups of a molecule are stable enough to determine the property, while scaffolds (environmental features) are irrelevant. 
Due to the instability and variability of environmental features, the distribution shifts in the graph can be further categorized into correlation shift and covariate shift: (1) Correlation shift. As illustrated in Figure \ref{fig: correlate shift}, environment features establish the spurious relation with labels. It means the GNN model learns a spurious relation between environment features and labels. Traditional GNNs assume that the training set encompasses all environment features present in the test set, but in reality, the assumption often gives rise to inconsistent statistical relationships between the training and test sets.
(2) Covariate shift. Environment features are very different between the training set and test set \cite{gui2022good, sui2024unleashing}. Considering an example in Figure \ref{fig: covariate shift},  when the test set contains new or unseen environmental features that are not accounted for in the training set, a covariate shift occurs. Covariate shift often happens in a situation where the test set has new environmental features, which do not appear in the training set. This phenomenon is frequently observed in scenarios where the training dataset is of insufficient quantity or variety.

\begin{figure}[]
    \centering
    \vspace{0.6cm} 
    \includegraphics[scale=0.6]{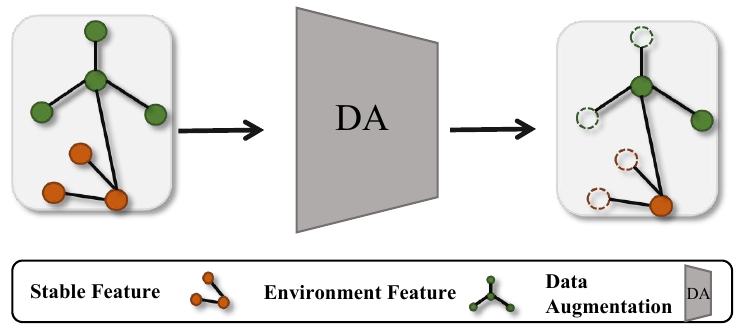}
    \caption{Traditional graph data augmentation. Traditional data augmentation strategies may destroy the structure of stable features.}
    \label{fig: traditional DA}
\end{figure}

 Currently, there are two primary approaches to addressing the OOD problem:  
 (causal) invariant learning and data augmentation \cite{kong2022robust, sui2024unleashing}. (i) Invariant learning approaches seek to identify a subset of graph features, referred to as the rationale (stable feature), that are most informative and predictive, thereby providing a robust explanation for the model's predictions. However, invariant learning relies on the assumption that the environment features present during training are identical to those encountered during testing. Therefore, it neglects the potential differences in environmental features between the two phases. In other words, graph invariant learning does not consider the covariate shift situation. (ii) Data augmentation approaches utilize the perturbation at different levels, such as node features, quantity of nodes, edges, or sub-graphs to tackle the OOD problem. However, graph data augmentation often fails to distinguish stable features, inadvertently introducing noise that disrupts the entire graph, ultimately destroying the features that are essential for robust prediction. The traditional graph data augmentation methods are defective in tackling the graph OOD problem in correlation shift distribution. As illustrated in Figure \ref{fig: traditional DA}, these methods have unlimited data augmentation strategies, which may destroy the structure of stable features, resulting in a suboptimal performance in the situation of covariate distribution shift.

The primary challenge lies in the fact that most existing methods overlook the covariate shift issue, neglecting its significant impact on model performance. A novel approach, adversarial invariant augmentation (AIA) \cite{sui2024unleashing} was proposed, which not only effectively distinguishes between stable features and environmental features but also specifically targets the environmental features for perturbation. Upon re-examining this method, we discover that AIA has an insufficient latent information utilization issue where the boundary between different types of features in the latent space is not sufficiently well-defined.

To further utilize the information in the latent space and mitigate the issue above, we utilize the manifold assumption \cite{van2020survey}, which posits that similar predictions from the network imply proximity in the manifold, to strengthen the performance of AIA. Therefore, we propose a novel method called MPAIACL, for \underline{M}ore \underline{P}owerful \underline{AIA} using \underline{C}ontrastive \underline{L}earning in this paper. We leverage contrastive learning to strengthen both the stable features generator and the adversarial augmenter. MPAIACL brings the embeddings of stable features closer together in latent space, while simultaneously distancing them from environment features, thereby enhancing the performance of the original model. We validated MPAIACL on various datasets, including molecule datasets like  Molbbbp, Molbace in OGB \cite{hu2020open}, and MolHiv in GOOD \cite{gui2022good}. Moreover, we validate the effectiveness of our approach across diverse datasets from different domains, varying in size and scaffold, using the same parameter settings as in \cite{sui2024unleashing}. Our experimental results demonstrate the superior generalization and effectiveness of our method. In summary, the main contributions of this paper are as follows:
 
 \begin{itemize}
     \item We uncover the under-explored representation information in AIA, which can be leveraged to further enhance the performance of the original model. 
     
     \item We adopt contrastive learning to strengthen the original AIA model, strengthening both the stable feature generator and the adversarial augmenter. 
     
     \item We provide a theoretical analysis of our approach, offering insights into its underlying mechanisms and effectiveness.
     
     \item Through extensive experiments and analysis, we demonstrate the generalization and effectiveness of MPAIACL.            
 \end{itemize}

After the introduction, we reviewed the related work, including graph contrastive learning and graph invariant learning in Section \ref{sec: related work}. The motivation of MPAIACL is illustrated in Section \ref{sec: motivation}. In Section \ref{sec: preliminaries}, we illustrate the preliminaries and the notations. We briefly introduce the GNNs used in graph classification and the related definitions of OOD problems. Then we provide the problem statement in this paper. In Section \ref{sec: methodology}, we explain the methodology of MPAIACL and conduct a theoretical analysis. In Section \ref{sec: experiment}, we provide the experiment analysis and the limitations of our method. Finally, we conclude our work in Section \ref{sec: conclusion}.

\section{Related Work} \label{sec: related work}
In this section, we provide a review of the related literature on graph contrastive learning and graph invariant learning. We introduce the related classical methods and then discuss the pros and cons of different methods.

\subsection{Graph Contrastive Learning}

Contrastive learning is a method that makes the representation of proper transformations of data agree with the positive one and be as far away from the negative one as possible. It is a traditional self-supervised learning method that has yielded impressive results in the field of computer vision \cite{ji2019invariant}. With the rapid growth of interest in graph contrastive learning, a multitude of methods have emerged in recent years \cite{shen2025heterophily, wang2025multi,  xu2025contrastive, yang2025dual}. Graph contrastive learning leverages multiple views of varying scales to enhance the embedding representation of similar instances. It can be broadly categorized into two paradigms: intra-scale and inter-scale contrast \cite{ju2024towards}.

Intra-scale contrast refers to the process of contrasting information within the same scale, such as at the local level \cite{liu2023b2, zhu2020deep}, contextual level \cite{liu2024multi, qiu2020gcc}, or graph level \cite{you2020graph, you2021graph}. (1) Local-level contrast focuses on learning node-level representations by comparing and aligning node-centric embeddings, thereby capturing fine-grained differences between individual nodes. GRACE \cite{zhu2020deep} generates two graph views through corruption and learns node representations by maximizing the consistency. GCA \cite{zhu2021graph} captures both topological and semantic aspects of the graph. B2-Sampling \cite{liu2023b2} learns to correct and refine the labels of error-prone negative pairs during training. (2) Context-level contrast refers to contrasting methods at the subgraph level. GCC \cite{qiu2020gcc} captures universal topological properties across multiple networks. MSSGCL \cite{liu2024multi} introduces a multi-scale subgraph contrastive learning approach that effectively captures fine-grained semantic information. (3) Graph-level contrast employs discrimination between graph representations. GraphCL \cite{you2020graph} proposes four distinct graph augmentation methods to generate varied views. JOAO \cite{you2021graph} simultaneously refines the graph augmentation selection and enhances the contrastive objectives. AD-GCL \cite{suresh2021adversarial} proactively identifies and minimizes redundant information. HGCL \cite{ju2023unsupervised} explores the hierarchical structural semantics of a graph.

Intra-scale contrast refers to the process of contrasting information across scales, encompassing various types of contrastive relationships, such as local-global contrast \cite{velivckovic2018deep}, local-context contrast \cite{jiao2020sub, mavromatis2020graph}, and context-global contrast \cite{cao2021bipartite, sun2021sugar}. (1) Local-global contrast seeks to capture the intricate relationships between local and global information, for maximizing the mutual information between these two scales. DGI \cite{velivckovic2018deep} aligning patch representations with high-level summaries of graphs. CGKS \cite{zhang2023contrastive} improves generalization ability and incorporates awareness of latent anatomies. (2) Local context contrast focuses on capturing the properties of subgraphs rather than the full graph. SUBG-CON \cite{jiao2020sub} exploits the correlation between central nodes and their associated subgraphs. GIC \cite{mavromatis2020graph} aims to further enrich graph representations by capturing cluster-level information content. HCHSM \cite{tu2023hierarchically} integrates multiple levels of intrinsic graph features to capture the hierarchical relationships within the graph. (3) Context-global contrast aims to strengthen the mutual information between subgraph representations and the overall graph representation. SUGAR \cite{sun2021sugar} reconstructs a sketched graph by identifying and extracting striking subgraphs. BiGI \cite{cao2021bipartite} encodes the global characteristics of bipartite graphs. MICRO-Graph \cite{zhang2024motif} extracts informative motifs and subsequently utilizes these learned motifs to guide the sampling of informative subgraphs for contrast. 

However, a significant limitation of most existing methods is that they overlook the importance of invariant graph features, which can also serve as a valuable source of contrastive information. In our method, we not only consider the stable features but also utilize the latent space information in a contrastive way.

\subsection{Graph Invariant Learning}

Graph invariant learning is a methodology that identifies and extracts the invariant properties of graphs, which are then leveraged to enhance the model's generalization capabilities. There also exists pioneering work \cite{wu2022handling} that leverages the invariant principle to identify stable properties to address the OOD generalization problem. For example, GREA \cite{liu2022graph} separates rationale and environment in latent spaces and performs representation learning on both real and augmented examples. CAL \cite{sui2022causal} identifies causal patterns in graph data and mitigates the confounding effects of shortcuts. FLAG \cite{kong2022robust} enhances node features with adversarial perturbations during training. AIA \cite{sui2024unleashing} increases the features of the environment and effectively addresses the problem of covariate distribution shift. Beyond the aforementioned methods, other generalization algorithms exist. IRM \cite{arjovsky2019invariant} estimates nonlinear, invariant, and causal predictors from multiple training environments. GroupDRO \cite{sagawadistributionally} learns models by minimizing the worst-case training loss across a set of predefined groups. GALA \cite{chen2024does} learns invariant graph representations under the guidance of an environment assistant model. However, most existing methods overlook the importance of tackling stable features and environment features from a contrastive perspective. Our method facilitates the convergence of the stable features and promotes the divergence of environmental features.

\begin{table}[ht]
\caption{Notations with the corresponding explanations.}
\label{table: notation}
\begin{tabular}{c|l}
\hline
\multicolumn{1}{c|}{\textbf{Notation}} & \multicolumn{1}{c}{\textbf{Explanation}}          \\ \hline
$\mathcal{G}$    & The graph data.                 \\
$\mathcal{V}$    & The node set of a graph.        \\
$\mathcal{X}$    & The feature set of $\mathcal{V}$ in a graph. \\
$\mathcal{A}$    & The adjacent matrix of a graph. \\
$\mathcal{Y}$    & The label of the graph.          \\
$\mathcal{D}$                                       & The dataset that includes various graph.            \\
$DS$   & The wasserstein distance.       \\
$\mathcal{L}$ & The loss value. \\
$\mathbb{E}(\cdot, \cdot)$  & The empirical risk.             \\
$\ell(\cdot, \cdot)$  & The loss function.              \\
$P (\cdot, \cdot)$ & The probability functions.       \\ 
$*^{tr}$                                    &The * of training set. \\
$*^{ts}$ & The * of test set.      \\
$*_{std}, *^{std}$    & The * of stable features. \\
$*_{DA}, *^{DA}$ &     The * of augmentation features. \\ 
$*_{env}, *^{env}$  & The * of environment features.\\

\hline

\end{tabular}
\end{table}

\begin{figure*}[]
    \flushleft 
    \subfigure[Feature distribution of AIA.]{
        \label{fig: AIA_tse}
        \includegraphics[scale = 0.22]{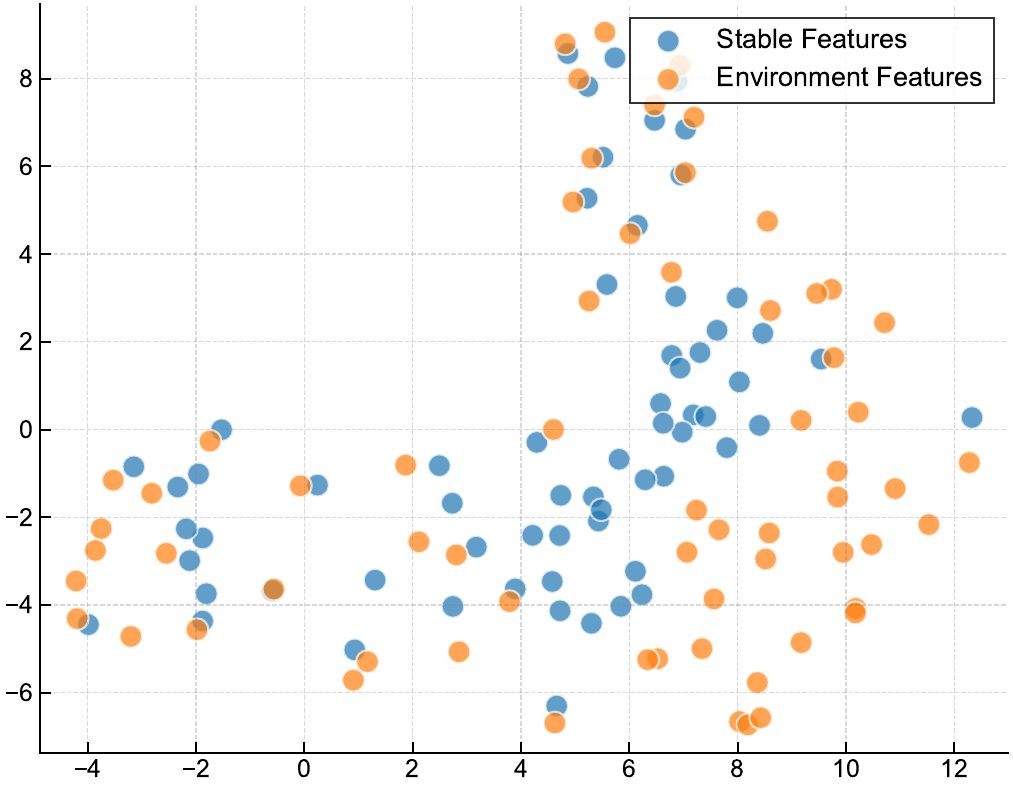}
    }
    \subfigure[Feature distribution of MPAIACL.]{
        \label{fig: MPAIACL_tse}
        \includegraphics[scale = 0.22]{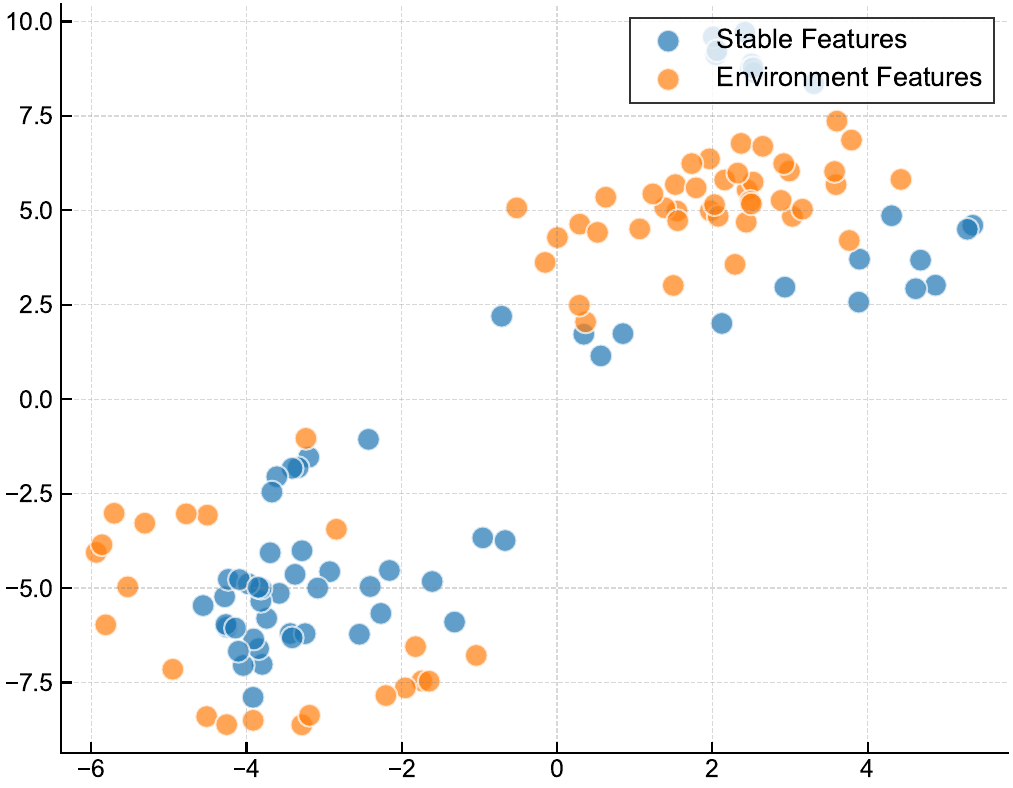}
    }\hspace{-0.2cm}
    \subfigure[Euclidean distance comparison.]{
        \label{fig: distance}
        \includegraphics[scale = 0.23]{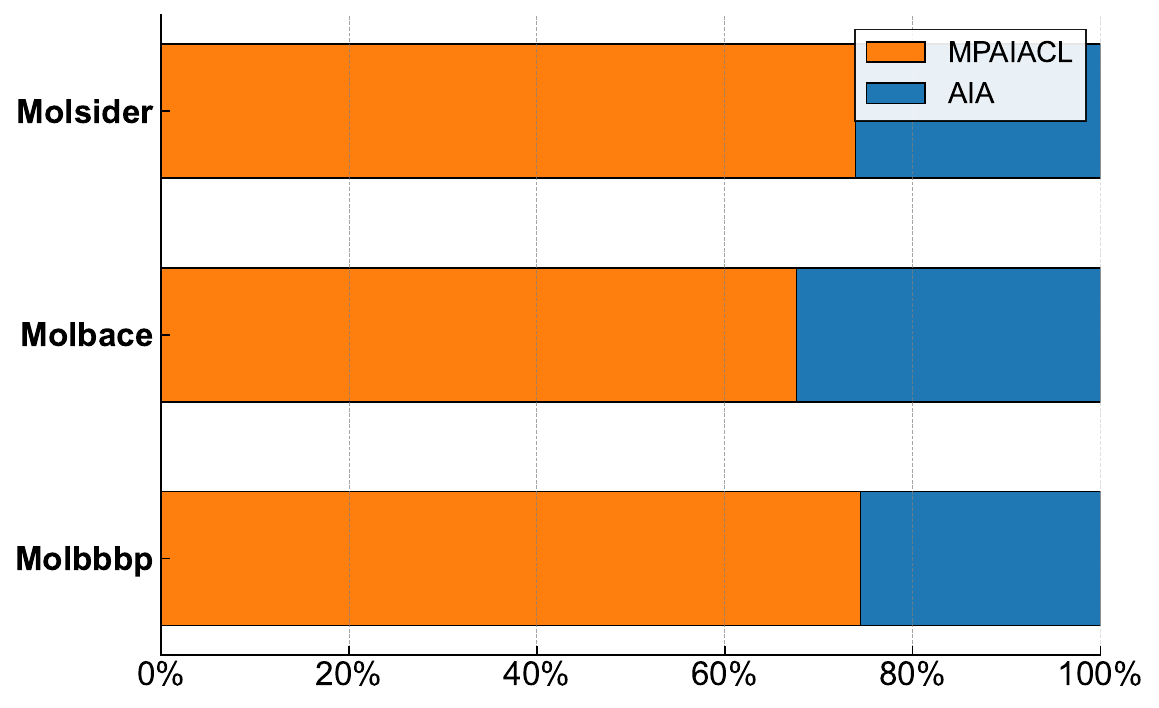}
    }\hspace{-0.2cm}
    \subfigure[Accuracy comparison.]{
        \includegraphics[scale = 0.23]{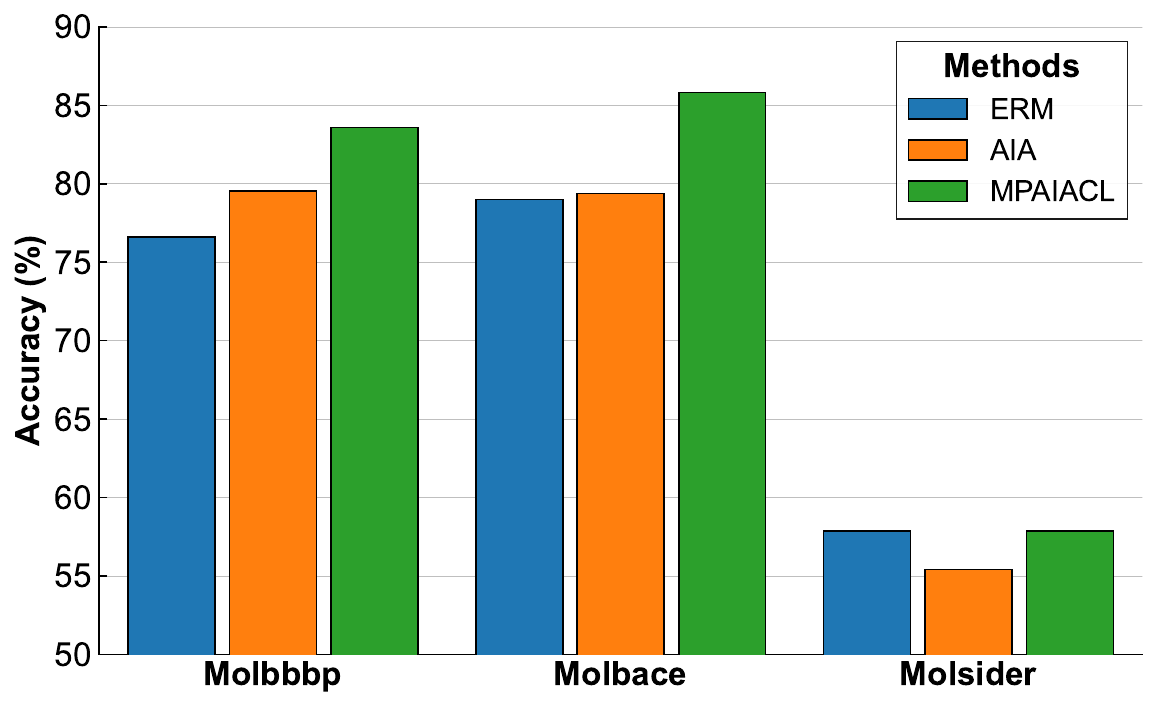}
        \label{fig: motivation_acc}
    }
    
    \caption{Result of the experiment in section \ref{sec: motivation}. 
    (a) and (b) visualize the stable feature and environment feature distribution in the latent space. 
    (c) Visualize the average Euclidean distance of the stable feature and environment feature between AIA and MPAIACL.
    (d) Visualize the accuracy comparison between different datasets and methods. 
    }
    \label{fig: motivation}
\end{figure*}

\section{Insufficient Latent Information Utilization Issue}
\label{sec: motivation}
This section introduces the insufficient latent information utilization issue and presents our approach to fully exploiting the potential of the latent space.

\subsection{Cause Analysis}
The insufficient latent information utilization issue in AIA refers to the phenomenon where the boundary between different types of features in the latent space is not sufficiently well-defined. We analyze this issue from two perspectives: the feature distribution in the latent space and the Euclidean distance between representations. 

From the perspective of feature distribution, we refer to Figure \ref{fig: AIA_tse}, which visualizes the distributions of stable and environmental features in the latent space. The visualization is conducted under a covariate shift in the Molbbbp size domain using AIA. As illustrated in Figure \ref{fig: AIA_tse}, the blue dots represent stable features, while the orange dots correspond to environment features. The boundary between these two types of features remains indistinct, with some features of different types even overlapping or connecting with each other.

From the perspective of Euclidean distance, Figure \ref{fig: distance} presents the average Euclidean distance ratio between stable and environmental features within the latent space. The visualization is obtained under covariate shift in the Molbbbp size domain using the AIA framework. As shown in Figure \ref{fig: distance}, AIA exhibits a significantly smaller distance proportion compared to MPAIACL, suggesting that the stable and environmental features extracted by AIA are more closely aligned in the latent space.

In summary, the core cause of the insufficient latent information utilization issue in AIA lies in the relatively large distance between the stable and environment features within the latent space.

\subsection{Solution}
In this subsection, we provide a step-by-step analysis of how our method addresses the insufficient latent information utilization issue in AIA.

Based on the above summary, we hypothesize that leveraging the mutually exclusive relationship between stable features and environment features could further enhance the model’s discriminative ability. Therefore, we utilize the manifold assumption \cite{van2020survey} and introduce the concept of contrastive learning to enable the model to acquire a stronger discriminative capability. Based on this idea, we propose MPAIACL for \underline{M}ore \underline{P}owerful \underline{A}dversarial \underline{I}nvariant \underline{A}ugmentation using \underline{C}ontrastive \underline{L}earning. To fully unleash the potential of the information embedded in the latent space, we employ contrastive learning to push apart the stable and environmental features within this space. At the same time, we leverage contrastive learning to enhance the representation of environment features.

From the perspective of feature distribution in the latent space, Figure \ref{fig: MPAIACL_tse} illustrates the distributions of stable and environment features under a covariate shift in the Molbbbp size domain using AIA. As shown in Figure \ref{fig: MPAIACL_tse}, MPAIACL exhibits a more distinct and well-separated boundary between stable and environmental features compared to AIA, indicating improved feature discrimination in the latent space.

From the perspective of Euclidean distance, as shown in Figure \ref{fig: distance}, the distance proportion of MPAIACL is significantly larger than that of AIA, indicating that the stable and environmental features extracted by MPAIACL are more distinct in the latent space.

Additionally, as shown in Figure \ref{fig: motivation_acc}, the results indicate a positive correlation between the distinctness of stable and environmental feature proportions and the model’s accuracy—that is, the greater the separation between these features, the higher the performance achieved.

\section{Preliminaries} \label{sec: preliminaries}

In this section, we present the preliminary concepts and definitions that are used in this work. We also introduce the notations used throughout the paper. Additionally, we provide a brief overview of the graph neural network (GNN) architecture employed for graph classification and define the distribution shift.
Finally, we present a detailed formulation of the problem statement, highlighting the key challenges and objectives that our proposed approach seeks to address.

In this paper, we focus on supervised graph classification and perform the task on many undirected graphs. An undirected graph is denoted as $\mathcal{G}$ ($\mathcal{V}$, $\mathcal{X}$, $\mathcal{A}$), where $\mathcal{V}$ denotes the node sets, $\mathcal{X}$ is a feature set, and $\mathcal{A}$ is an adjacent matrix. $\mathcal{Y}$ is the label set, including the labels of each graph. We also use $\mathcal{D}$ = $\{(\mathcal{G}, \mathcal{Y})\}$ to denote datasets, where $\mathcal{D}_{tr}$ = $\{(\mathcal{G}^{tr}, \mathcal{Y}^{tr})\}$ denotes a training set, and $\mathcal{D}_{ts}$ = $\{(\mathcal{G}^{ts}, \mathcal{Y}^{ts})\}$ denotes a test set.  There are two core concepts used in the entire paper, which are i) Stable features, which capture the underlying patterns of the entire graph. The relationship between stable features and labels can be considered invariant. ii) Environment features, which are subject to variation for the label \cite{sui2024unleashing}. The notations used throughout this paper are summarized in Table \ref{table: notation}.

\subsection{Graph Neural Networks}

GNNs are a class of deep learning architectures specifically designed to process and analyze graph-structured data. They have demonstrated remarkable effectiveness across a wide range of tasks, including node-level, edge-level, and graph-level classification \cite{wang2022powerful}. In this paper, we focus on the supervised graph classification task. We provide an overview of how GNNs operate for graph classification tasks below:
\begin{equation}
\begin{aligned}
\mathbf{h}_v^{(k)} &= \text{Aggregate}\left(\mathbf{h}_v^{(k-1)}, \left\{\mathbf{h}_u^{(k-1)} \mid u \in \mathcal{N}(v)\right\}\right). \\
h_\mathcal{G} &= \text{READOUT}\left(\{h_v^{(K)}: v \in \mathcal{G}\}\right).
\end{aligned}
\end{equation}
Here, $h^{(k-1)}_v$ represents the embedding of the node $v$ at the ($k$-1)-th layer, $\mathcal{N}(v)$ denotes the set of neighbors of node $v$. Aggregate is a function that aggregates the features of the node with its neighbors. 

In summary, Graph Neural Networks (GNNs) generally consist of three main steps: message passing, aggregation, and readout. Each node $v$ is initially represented by a feature vector $h_v^{(0)}$. (1) During message passing, nodes exchange information with their neighbors to compute messages. (2) In the aggregation step, each node aggregates the messages from its neighbors and updates its own representation $h_v^{(l+1)}$. (3) Finally, in the readout phase, graph-level representations are derived for downstream tasks. Subsequently, the readout process yields the graph-level representation $h_\mathcal{G}$.

\subsection{Definition}

From the perspective of invariant learning and stable learning \cite{liu2022graph, wu2022discovering}, a fundamental assumption is that there exist stable features that are determinative of the label in classification tasks \cite{sui2022causal, wu2022discovering, yang2022learning}. 

The relationship between stable features and labels provides a foundation for tackling OOD generalization \cite{sui2024unleashing}. On the contrary, the other scaffold structure is generally irrelevant to the properties, which can be considered as environmental features \cite{hu2020open, yang2022learning}. The environmental features don't have a direct effect on labels. Due to the inherent imprecision in data collection and the limitations of environmental features, inconsistencies inevitably arise between the training set and test set, giving rise to two primary OOD challenges: correlation shift and covariate shift \cite{ gui2022good}.

\begin{definition} [Correlation shift]
    \rm  The training set has a delusive correlation between the data and labels, which is not established in the test set. The graph correlation shift can be denoted as:
    \begin{equation}
        P_{tr}(\mathcal{G} | \mathcal{Y}) \neq P_{ts}(\mathcal{G} | \mathcal{Y}), P_{tr}(\mathcal{G}) = P_{ts}(\mathcal{G}).
    \end{equation} 

    Equation (2) indicates the presence of false statistical correlations in the training data, which may not hold true in the testing data. Besides, $P_{\text{tr}}(\mathcal{G})$ = $P_{\text{ts}}(\mathcal{G})$ indicates that the overall distribution of graph structures in the training and testing sets is the same. Notice that $P$ denotes the probability distribution function.
\end{definition}

\begin{definition} [Covariate shift] 
   \rm  The test set has new environment features, which do not appear in the training set. It may happen due to an insufficient quantity or variety of datasets. The graph covariate shift can be denoted as $P_{tr}(\mathcal{G} | \mathcal{Y}) = P_{ts}(\mathcal{G} | \mathcal{Y}), P_{tr}(\mathcal{G}) \neq P_{ts}(\mathcal{G})$. The equality $P_{\text{tr}}(\mathcal{G} \mid \mathcal{Y}) = P_{\text{ts}}(\mathcal{G} \mid \mathcal{Y})$ indicates that the statistical correlation between the graph structures and the labels is consistent across the training and testing sets, suggesting that the learned relationships can generalize from the training set to the test set. In contrast, $P_{\text{tr}}(\mathcal{G}) \neq P_{\text{ts}}(\mathcal{G})$ reveals that there are discrepancies in the graph structures between the training and test sets, which can be attributed to differences in environmental features. We can measure Graph Covariate Shift (GCS) as \cite{sui2024unleashing}:

   \begin{equation}
       GCS(P_{te}, P_{ts}) = \frac{1}{2}\int_{S}|P_{tr} - P_{ts}| dg.
   \end{equation}

\end{definition}
where $S = \{P_{tr}(\mathcal{G}) \cdot P_{ts}(\mathcal{G}) = 0\}$, which means the features (environmental features) don't overlap in both sets. GCS is used to measure the difference in distribution between the training set and the test set. The covariance shift is a pervasive issue in real-world applications \cite{sui2024unleashing}.

\subsection{Problem Statement}

The objective of graph classification under covariate distribution shift is to identify and disentangle stable features from environmental features. 

Subsequently, the GNN encoder is trained in a manner analogous to traditional graph classification tasks. The problem can be defined as \cite{sui2024unleashing}:
\begin{equation}
    f^* = \arg \min_f \sup_{e \in \mathcal{E}_{\text{te}}} \mathbb{E}_e[\ell(f(\mathcal{G}), y)].
\end{equation}

$\mathbb{E}_e[\ell(f(\mathcal{G}), y)]$ denotes the empirical risk of the environment $e$, $\mathcal{E}_{\text{te}}$ is the test environment, $\ell(\cdot, \cdot)$ is the loss function, and $f$ is the GNN encoder. The goal of our paper is to train a model $f$ that can minimize the difference of environment between the training and the test set to alleviate the covariate shift. In this paper, we adopt contrastive learning \cite{jiao2020sub} to further disentangle stable features from environmental features, therefore enhancing the capabilities of the original model.

\begin{figure*}[h]
    \centering
    \vspace{0.6cm} 
    \includegraphics[scale=0.55]{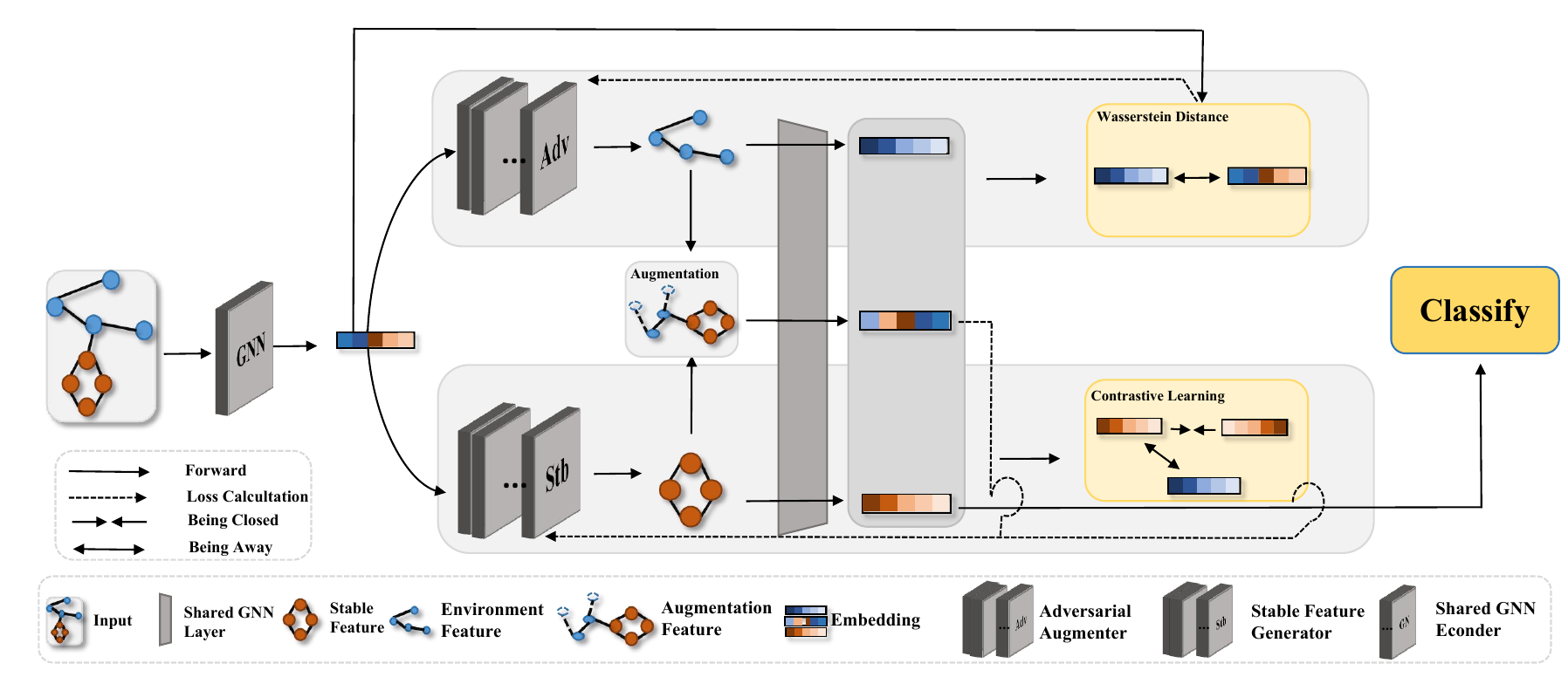}  
     \caption{The overview of MPAIACL. The explanation of the notation used in the figure is located at the bottom and the bottom left. The training process consists of two distinct phases: (1) Strengthen the Stable Feature Generator (SFG), and (2) Strengthen the Adversarial Augmenter (AA). Initially, the input graph is processed through a shared GNN encoder, which generates the graph embeddings. 
     In phrase (1), SFG generates the stable features. Then, the stable feature $\mathcal{F}_s$ would make a data augmentation with the environment feature $\mathcal{F}_e$, which is generated by AA to obtain the augmentation feature $\mathcal{F}_a$.  Subsequently, MPAIACL utilizes contrastive learning to pushing away $\mathcal{F}_s$ from $\mathcal{F}_e$. Finally, the results of the contrastive and $\mathcal{F}_a$ are used to strengthen the SFG. In phrase (2), $\mathcal{F}_e$ uses Wasserstein distance to push away from $\mathcal{F}_s$ and the original graph embedding. The result of the Wasserstein distance is used to strengthen AA.}
     
    \label{fig: MPALACL}
\end{figure*}

\section{Methodology} \label{sec: methodology}

In this section, we first review the original adversarial invariant augmentation (AIA) model \cite{sui2024unleashing}. Then, we present our contrast approach, which involves strengthening the stable feature generator and the adversarial augmenter. The overall architecture of our method is illustrated in Figure \ref{fig: MPALACL}.

Initially, the input data is encoded by a shared GNN, then processed by the adversarial augmenter and the stable feature generator, respectively. Subsequently, we extract the stable feature and environment feature, which are then combined to generate the data augmentation version of the input data. Finally, we leverage the embeddings of the stable feature and environment features to optimize the model parameters. For the adversarial augmenter, we employ the triplet loss \cite{schroff2015facenet} as a new metric to measure the discrepancy between stable features and environment features. This enables us to push the environment features away from the stable features. For the stable feature generator, we employ InfoNCE \cite{chen2020simple}, a contrastive learning loss, to effectively harness the vector information present in the latent space. 

\subsection{Original Model Introduction}

In this subsection, we introduce the original model, namely AIA \cite{sui2024unleashing}, elaborating on its architecture and the underlying principle designed to maintain feature consistency.

\subsubsection{Model Architechure} \label{sec: Model Architechure}
Adversarial invariant augmentation (AIA) is a model that utilizes data augmentation to tackle the covariate distribution shift problem. It consists of two primary components: a stable feature generator and an adversarial augmenter: (1) The stable feature generator is designed to produce stable features of $\mathcal{G}$ that satisfy the stable feature consistency;  (2) The adversarial augmenter aims to generate the environment features of $\mathcal{G}$, which adheres to environmental feature discrepancy. The environment features are used to strengthen the capability of distinguishing the stable features of the model. Then, AIA combines stable features with disturbed environment features to augment the graph $\mathcal{G'}$, which is named the data augmentation (DA) graph in the following section.

The stable feature generator and adversarial augmenter share the same architecture and are parameterized by $\theta_1$ and $\theta_2$, respectively. Given an input graph $\mathcal{G}$ with $n$ nodes, the mask generation network first computes node representations using a GNN encoder $\tilde{h}(\cdot)$. To assess the importance of nodes and edges, it utilizes two MLP layers, $\text{MLP}_1(\cdot)$ and $\text{MLP}_2(\cdot)$, to generate a soft node mask matrix $M_x \in \mathbb{R}^{n \times 1}$ and an edge mask matrix $M_a \in \mathbb{R}^{n \times n}$, respectively. After masking the features, the remaining features are represented as stable or environmental features. The mask generation process can be summarized as follows:
{\small
\[
Z = \tilde{h}(g), \quad M_{x_i} = \sigma(\text{MLP}_1(h_i)), \quad M_{a_{ij}} = \sigma(\text{MLP}_2([z_i, z_j])).
\]
}

\subsubsection{Proposed Principle}

To maintain consistency between the stable feature and the original graph, and to address the discrepancy between the environment features and the original graph, AIA proposed two principles: (1) the environmental feature discrepancy principle and (2) the stable feature consistency principle.

\begin{principle} [Environmental feature discrepancy]
    \rm  Let \textit{D-Aug}$\{\cdot\}$ denote data augmentation and $P$ denote the data distribution. For environment features, given a graph $\mathcal{G}$, the augmented graph $\mathcal{G}'$ = $DAug\{\mathcal{G}\}$ should satisfy \textit{GCS} = $\{P(\mathcal{G}), P(\mathcal{G}')\} $ $\rightarrow$ 1.
\end{principle}

\textit{GCS} = $\{P(\mathcal{G}), P(\mathcal{G}')\}$ $\rightarrow$ 1 means the environment features should keep away from the original distribution. From the viewpoint of data distribution, environment features should remain inconsistent with the original graph. In this paper, we also strengthen the discrepancy principle by keeping the environment features away from the stable feature, the original graph, and the DA graph. 

\begin{principle} [Stable feature consistency]
  \rm   Let \textit{DAug}$\{\cdot\}$ denote data augmentation, $\mathbb{E}$ denote the empirical risk, and $\mathcal{G}_{std}\{\mathcal{A}_{std}, \mathcal{X}_{std}\}$ denote a stable feature set. Then, the augmented graph $\mathcal{G}'_{std} \{\mathcal{A}'_{std}, \mathcal{X}'_{std}\}$ = \textit{DAug}$\{\mathcal{G}\}$ should meet $\mathbb{E}\left[\|A_{\text{std}} - \mathcal{A}'_{std}\|^2_F\right] \to 0 \quad \text{and} \quad \mathbb{E}\left[\|X_{\text{std}} - \mathcal{X}'_{std}\|^2_F\right] \to 0$.
\end{principle}

The above formula implies that the generated stable features should preserve the consistency of the original graph. In this paper, we further strengthen the consistency between the stable features and the DA graph, thereby enhancing the model's generalization ability.

Finally, AIA utilizes cross-entropy to train the whole model \cite{sui2024unleashing}. AIA uses a very imaginative method to tackle the covariate distribution shift issue in graphs. However, the original approach overlooked the intrinsic correlation of the vector itself in the latent space, relying solely on label information. In contrast, we strengthen the AIA with contrastive learning to unleash the power of the vector itself.

\subsection{MPAIACL} 

In this subsection, we introduce the core method. The overall architecture of our method is illustrated in Figure \ref{fig: MPALACL}. The optimization of AIA and MPAIACL formulated as a two-player min-max game between the Stable Feature Generator (SFG) and the Adversarial Augmenter (AA). SFG aims to distinguish stable features from the graph, while the AA attempts to produce environment features that are indistinguishable from the graph. During training, the two networks are optimized alternately: In each training iteration, the SFG is first optimized. Upon completion of SFG optimization, the AA is subsequently updated. The completion of AA optimization marks the end of a single training epoch. This iterative procedure is repeated until either model convergence is achieved or the maximum number of training iterations is reached. An illustration of the training process is provided in Figure \ref{fig: MPALACL} and its corresponding caption.

\subsubsection{Strengthen the Stable Feature Generator}

Our work is motivated by the success of contrastive learning with data augmentation in self-supervised learning, which has also been shown to be effective in graph-based applications \cite{ju2024towards}. However, there are various contrastive losses. For example, triplet loss enforces a margin between the anchor and positive samples, and the anchor and negative samples, such that the distance between the anchor and positive is always smaller. Information noise contrastive estimation (InfoNCE) \cite{chen2020simple} can learn representations of data such that positive examples are closer to each other in the feature space than negative examples (e.g., different images). Normalized temperature-scaled cross-entropy loss (NT-Xent) can build on the InfoNCE loss but introduces specific normalization and temperature scaling to improve training stability and performance. In this work, we adopt the widely used InfoNCE loss as our contrastive loss. It should be noted that any contrastive loss satisfying the above requirements can be incorporated into the proposed algorithm. In this work, we implement the algorithm using the InfoNCE loss.

The original model utilizes the cross-entropy with label information to train. Here, we utilize InfoNCE \cite{chen2020simple} to unleash the power of the vector itself. InfoNCE is defined as follows:
\begin{equation} \label{equation: contrastive loss}
\scalebox{0.8}{
 $\displaystyle
\mathcal{L}_{\text{InfoNCE}} = -\frac{1}{N} \sum_{i=1}^{N} \log \frac{\exp(\operatorname{sim}(h^i_s, h^i_{DA}) / \tau)}{\exp(\operatorname{sim}(h^i_s, h^i_{DA}) / \tau) + \sum_{\substack{j=1 }}^{N} \exp(\operatorname{sim}(h^i_s, h^i_e) / \tau)}
$}
\end{equation}

Among the formulas, $h$ denotes the hidden representation of $\mathcal{G}$ in the latent space. $h^i_{s}$ is the i-th graph $\mathcal{G}$ only with stable features. $h^i_{e}$ is the i-th graph that only has environment features. 
$h^i_{DA}$ only with data augmentation features.
$N$ denotes the total number of samples. $\tau$ is the temperature coefficient. We use the InfoNCE to bring the augmented graph closer to the stable features while pushing it further away from the environment feature. The intuition is that the similar predictions of neural networks indicate the close proximity in the manifold \cite{van2020survey}. The InfoNCE loss encourages the representation of $h^i_{s}$ to be closer to $h^i_{DA}$, while pushing $h^i_{e}$ away from $h^i_{s}$. This contrastive learning process enhances the model's ability to distinguish the stable feature and the environmental feature. However, relying solely on InfoNCE would result in the vectors moving towards or away from each other without any constraints, leading to uncontrolled and chaotic behavior. To prevent the vectors from moving in an uncontrolled way, we also incorporate label information serving as a ground truth boundary to guide the optimization process. The formula is defined as follows:

\begin{equation}
    \mathcal{L}^{\text{std}}_{\text{Reg}} = -\left( \sum_{i} \mathcal{Y}_i \log (\text{Pred}^{\text{std}}_i) + \sum_{i} \mathcal{Y}_i \log (\text{Pred}^{\text{DA}}_i) \right).
\end{equation}

Here, $\mathcal{Y}_i$ is the label of each graph. $Pred$ denotes the prediction of each graph.  $\mathcal{L}^{\text{std}}_{\text{Reg}}$ is a supervised loss that acts as a constraint that anchors feature vectors to task-relevant representations. While $\mathcal{L}_{\text{InfoNCE}}$ structures the relative positions of vectors, $\mathcal{L}^{\text{std}}_{\text{Reg}}$ prevents them from drifting uncontrollably in the feature space, ensuring both stability and discriminative power. This joint optimization balances task alignment with contrastive separation.

Finally, we have a new optimization target:
\begin{equation} \label{equation: obj of stable generator}
    min \{\mathcal{L}_{std} = \mathbb{E}\,[\mathcal{L}^{\text{std}}_{\text{Reg}} + \lambda \mathcal{L}_{\text{InfoNCE}}]   \}.
\end{equation}

The formula implies that the optimal stable feature generator is obtained by minimizing the joint empirical risk of the supervised learning and the contrastive loss term. Note that $\lambda$ is a hyperparameter. Additionally, the Formula (\ref{equation: obj of stable generator}) not only enhances the capability of the model to distinguish the stable feature, but also prevents the representation vector from moving away without any constraint.

\begin{table*}[t]
\caption{Statistics of graph classification on the molecular datasets in the covariant shift.}
\small
\renewcommand\arraystretch{1.3}
\setlength{\tabcolsep}{5pt}
\scalebox{0.8}{
\begin{tabular}{cccccccccccccccc}
\hline
\multicolumn{2}{c}{\textbf{Dataset}}         & \multicolumn{2}{c}{\textbf{Molbace}} & \multicolumn{2}{c}{\textbf{Molbbbp}} & \multicolumn{2}{c}{\textbf{MolHiv}} & \multicolumn{2}{c}{\textbf{Molsider}} & \multicolumn{2}{c}{\textbf{Moltox21}} & \multicolumn{2}{c}{\textbf{Moltoxcast}} &\multicolumn{2}{c}{\textbf{Molclintox}} \\ \hline
\multicolumn{2}{c}{\textbf{Covariate  shift}} & \textbf{size}       & \textbf{scaffold}       &  \textbf{size}       & \textbf{scaffold}      & \textbf{size}       & \textbf{scaffold} &\textbf{size}       & \textbf{scaffold}&\textbf{size}       & \textbf{scaffold}&\textbf{size}       & \textbf{scaffold}&\textbf{size}       & \textbf{scaffold}     \\ \hline
\multirow{3}{*}{Train} & Graph \#    & 1211  & 1210  & 1633  & 1631  & 26169 & 24682 & 1143 & 1141 & 6265 & 6264 & 6862 & 6860 & 1183 & 1181 \\
                       & Ave node \# & 36.66 & 33.60 & 27.02 & 22.49 & 27.87 & 26.25 & 39.47 & 29.97 & 21.31 &   16.54 & 21.56 & 16.68 & 29.90 & 25.52 \\
                       & Ave edge \# & 79.05 & 72.59 & 58.71 & 48.43 & 60.20 & 56.68 & 83.47 & 62.81 & 44.72 & 33.74 & 44.66 & 33.54 & 64.17 & 54.10\\ \hline
\multirow{3}{*}{Valid} & Graph \#    & 151   & 151   & 203   & 204   & 2773  & 4113 & 142 & 143 & 783 & 783 & 857 & 858 & 147 & 148 \\
                       & Ave node \# & 23.69 & 37.23 & 12.06 & 33.20 & 15.55 & 24.95 & 10.52 & 43.24 & 7.60 & 26.76 & 7.71 & 26.17 & 11.50 & 32.75\\
                       & Ave edge \# & 52.17 & 81.29 & 24.27 & 71.81 & 32.77 & 54.53 & 20.09 & 91.84 & 14.06 & 53.13 &  14.10 & 56.09 & 22.78 & 71.36\\ \hline
\multirow{3}{*}{Test}  & Graph \#    & 151   & 152   & 203   & 204   & 3961  & 4108 & 142 & 143 & 783 & 784 & 857 & 858 & 147 & 148\\
                       & Ave node \# & 52.17 & 75.10 & 12.26 & 27.51 & 12.09 & 19.76  & 9.81 & 53.27 & 7.59 & 26.59 & 7.57 & 28.18 & 10.67 & 24.61  \\
                       & Ave edge \# & 52.47 & 34.82 & 24.87 & 59.75 & 24.87 & 40.58 & 18.60 & 112.65 & 14.05 & 57.77 & 13.72 & 60.70 & 21.14 & 53.44\\ \hline
\multicolumn{2}{c}{Class \#}         & 2     & 2     & 2     & 2     & 2     & 2 & 2 & 2 & 2 & 2 & 2 & 2 & 2 & 2\\ \hline
\end{tabular}
}
\label{table:statis info of covariant}
\end{table*}

\begin{table}[]
\caption{Statistics of graph classification on the GOOD datasets in covariate shift.}
\small
\renewcommand\arraystretch{1.2}
\setlength{\tabcolsep}{5pt}
\scalebox{0.8}{
\begin{tabular}{ccccccc}
\hline
\multicolumn{2}{c}{\textbf{Dataset}}           & \multicolumn{2}{c}{\textbf{MolHiv}} & \multicolumn{2}{c}{\textbf{Motif}} & \textbf{CMNIST}  \\ \hline
\multicolumn{2}{c}{\textbf{Covariate shift}} & \textbf{size}       & \textbf{scaffold}      & \textbf{size}       & \textbf{scaffold}      & \textbf{color}   \\ \hline
\multirow{3}{*}{Train}  & Graph \#     & 26169      & 24682         & 18000        & 18000      & 42000   \\
                        & Ave node \#  & 27.86      & 26.25         & 16.92        & 17.06      & 75.0    \\
                        & Ave edge \#  & 60.20      & 56.68         & 43.56       & 48.89      & 1393.15 \\ \hline
\multirow{3}{*}{Valid}  & Graph \#     & 2773       & 4113          & 3000         & 3000       & 7000   \\
                        & Ave node \#  & 15.54      & 24.94         & 39.22        & 15.82      & 75.0    \\
                        & Ave edge \#  & 32.77      & 54.53         & 107.03       & 33.00      & 1391.20 \\ \hline
\multirow{3}{*}{Test}   & Graph \#     & 3961      & 4108         & 3000         & 3000       & 7000   \\
                        & Ave node \#  & 12.09      & 19.76         & 87.18        & 14.96      & 75.0    \\
                        & Ave edge \#  & 24.86      & 40.57         & 239.64       & 31.54      & 1394.33 \\ \hline
\multicolumn{2}{c}{Class \#}           & 2          & 2             & 3            & 3          & 10      \\ \hline
\end{tabular}
}

\label{table:statis info of covariate of GOOD}
\end{table}

\begin{table}[]
\caption{Statistics of graph classification on the GOOD datasets in correlation shift.}
\small
\renewcommand\arraystretch{1.2}
\setlength{\tabcolsep}{5pt}
\scalebox{0.8}{
\begin{tabular}{ccccccc}
\hline
\multicolumn{2}{c}{\textbf{Dataset}}           & \multicolumn{2}{c}{\textbf{MolHiv}} & \multicolumn{2}{c}{\textbf{Motif}} &  \\ \hline
\multicolumn{2}{c}{\textbf{Correlation shift}} & \textbf{size}       & \textbf{scaffold}      & \textbf{size}       & \textbf{scaffold}         \\ \hline
\multirow{3}{*}{Train}  & Graph \#     & 14454      & 15209         & 12600        & 12600         \\
                        & Ave node \#  & 31.17      & 24.64         & 51.77        & 16.90         \\
                        & Ave edge \#  & 67.46      & 53.27         & 141.83       & 48.47       \\ \hline
\multirow{3}{*}{Valid}  & Graph \#     & 9956       & 9365          & 6000         & 6000         \\
                        & Ave node \#  & 20.06      & 26.35         & 51.47        & 17.03          \\
                        & Ave edge \#  & 42.89      & 56.58         & 140.20       & 48.91       \\ \hline
\multirow{3}{*}{Test}   & Graph \#     & 10525      & 10037         & 6000         & 6000       \\
                        & Ave node \#  & 19.39      & 26.64         & 51.60        & 17.01          \\
                        & Ave edge \#  & 41.42      & 57.21         & 141.51       & 48.69       \\ \hline
\multicolumn{2}{c}{Class \#}           & 2          & 2             & 3            & 3               \\ \hline
\end{tabular}
}

\label{table:statis info of correlation}
\end{table}

\subsubsection{Strengthen the Adversarial Augmenter}

\begin{figure}[]
    \centering
    \vspace{0.6cm} 
    \includegraphics[scale=0.35]{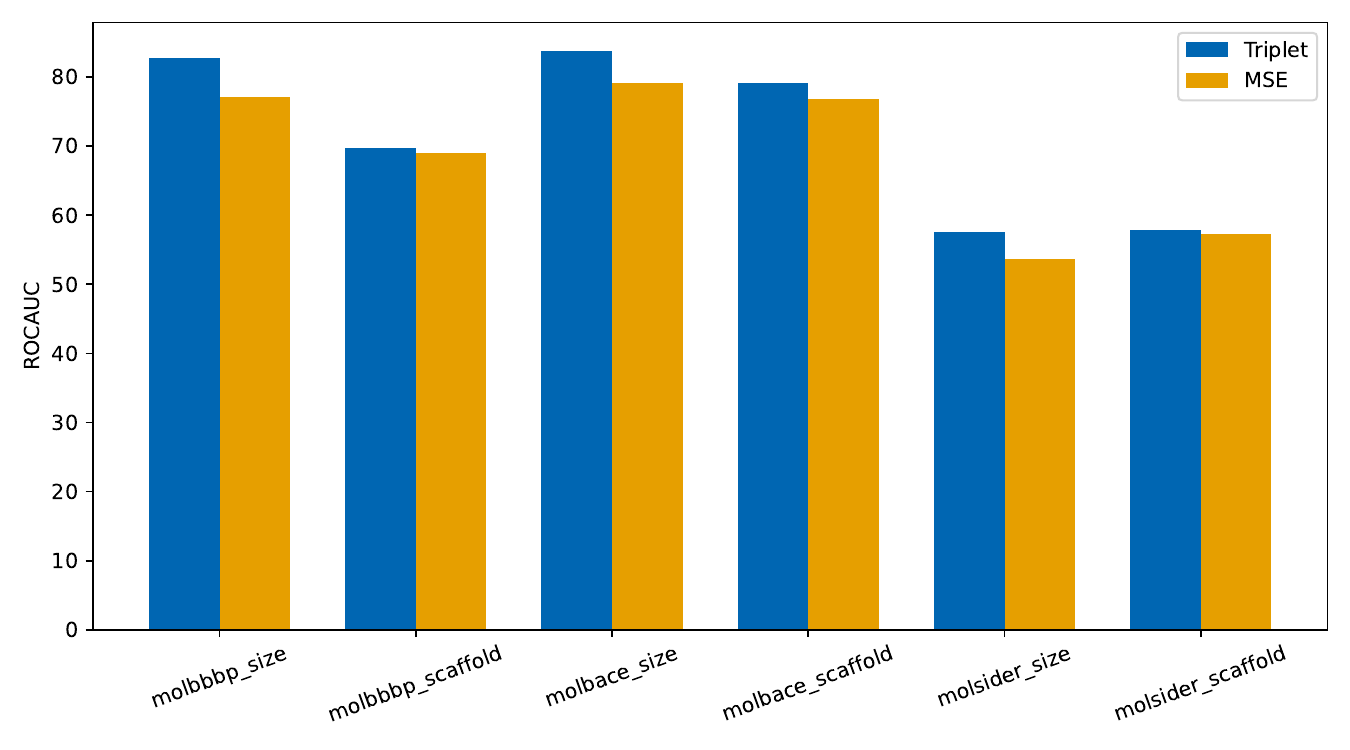}
    \caption{The performance compared between MSE and Triplet loss in three different datasets.}
    \label{fig: TripletVsMSE}
\end{figure}

To strengthen the adversarial augmenter, we considered utilizing contrastive learning to make environmental features different from the stable features. However, in this case, we lack a clear boundary to constrain the movement of the environmental features. Unlike the stable features, we do not have access to label information that can guide what constitutes an environmental feature. Therefore, we cannot utilize contrastive loss to guide the environment features toward their ground truth representations. We follow the method, which is Wasserstein distance \cite{arjovsky2017wasserstein} used in the original model as the distance metric to evaluate the distance between perturbation and the stable features. 

In this paper, we further increase the discrepancy between stable features and environment features by the vector information in the latent space using triplet loss \cite{schroff2015facenet}. We employ it instead of the MSE, as the latter only minimizes point-wise Euclidean distances and therefore fails to capture the relative geometric relationships among samples in the latent space. In contrast, the triplet loss introduces a margin-based constraint between positive and negative pairs, simultaneously reducing intra-class distances while enlarging inter-class distances. As illustrated in Figure \ref{fig: TripletVsMSE}, Triplet is better than MSE as distance metric. We evaluate the  Wasserstein distance between the original graph and the environmental features. Therefore, we have a new distance metric formula below:
\begin{equation}
\scalebox{1}{$
\begin{aligned}
\mathcal{L}_{\text{triplet}} &= \frac{1}{N} \sum_{i=1}^{N} \max \left( 0, \, d(\mathbf{h}_o^i, \mathbf{h}^i_{o^\prime}) - d(\mathbf{h}_o^i, \mathbf{h}^i_e + \alpha \right). \\
\end{aligned}
$}
\end{equation}
Here, $\mathbf{h}_o^i$ indicates the i-th representation of the original graph; $\mathbf{h}^i_{o^\prime} = \text{Dropout}(\mathbf{h}_o^i)$ indicates the dropout $\mathbf{h}_o^i$; $\mathbf{h}^i_e$ indicates i-th representation of the environment feature; $d(\mathbf{x}, \mathbf{y}) = \lVert \mathbf{x} - \mathbf{y} \rVert_2$ indicates the L2 norm; and $\alpha$ indicates the margin in triplet loss. Therefore, we have a brand-new optimization object based on AIA below:
\begin{equation} \label{equation: obj of Adversarial Augmenter}
    max\{\mathcal{L}_{adv} = \mathbb{E}[\mathcal{L}_s - \alpha \mathcal{L}_{\text{triplet}} - \gamma \mathcal{L}^{env}_{reg}]\}.
\end{equation}

Here, $\mathcal{L}_s =  -\left( \sum_{i} \mathcal{Y}_i \log (\text{Pred}^{\text{env}}_i) \right)$. Eq. \ref{equation: obj of Adversarial Augmenter} means we should maximize the empirical risk of the Wasserstein distance, label information, and the regularization term. $\alpha$ and $\gamma$ are the hyperparameters. The details of the regularization term $\mathcal{L}^{env}_{reg}$ can be referred to AIA \cite{sui2024unleashing}. Formula (\ref{equation: obj of Adversarial Augmenter}) maximizes the distances between $h_{\text{env}}$ and the original graph, while maintaining the original distribution of the dataset. To be more specific, the optimization objective in Eq. (9) is designed to construct challenging training environments by maximizing the supervised loss while simultaneously constraining the representation structure and environmental stability. Specifically, maximizing the supervised loss encourages the attacker to generate difficult environments, whereas the triplet loss preserves semantic consistency in the representation space. In addition, the environment regularization term prevents degenerate environment partitioning and ensures statistical validity. This adversarial mechanism ultimately promotes the learning of robust and invariant representations under distribution shifts. During the training process, $\mathcal{L}_{adv}$ exhibits an upward trend.

\subsection{Theoretical Analysis}

In this subsection, we present a theoretical analysis of our approach, elucidating the stable features that motivate our adoption of contrastive learning to fully exploit the potential of the latent space, and explaining why this strategy is effective in achieving the objectives.

We first define $h_s$, $h_{DA}$, and $h_e$, which denote the stable features, data augmentation features, and environment features, respectively. To enhance the stable feature generator, we employ contrastive learning, formulated as:
\begin{equation*}
\scalebox{0.8}{$\displaystyle
\mathcal{L}_{\text{InfoNCE}} = -\frac{1}{N} \sum_{i=1}^{N} \log \frac{\exp(\operatorname{sim}(h^i_s, h^i_{DA}) / \tau)}{\exp(\operatorname{sim}(h^i_s, h^i_{DA}) / \tau) + \sum_{\substack{j=1 }}^{N} \exp(\operatorname{sim}(h^i_s, h^i_e) / \tau)}
$}
\end{equation*}, which yields the enhanced stable features $h_s'$ and the augmented features $h_{DA}'$, respectively.
After applying contrastive learning, we obtain
\[
\|h_s' - h_e\| \gg \|h_s - h_e\|.
\]
These relationships indicate that, following contrastive learning, the strengthened stable features become more distinct from the environment features in the latent space. This separation enhances the model's ability to capture domain-invariant representations, thereby improving its overall discriminative performance.

To enhance the adversarial augmenter, we employ the Wasserstein distance defined as
\begin{equation*}
\scalebox{1}{$
\begin{aligned}
\mathcal{L}_{\text{triplet}} &= \frac{1}{N} \sum_{i=1}^{N} \max \left( 0, \, d(\mathbf{h}_o^i, \mathbf{h}^i_{o^\prime}) - d(\mathbf{h}_o^i, \mathbf{h}^i_e + \alpha \right), \\
\end{aligned}
$}
\end{equation*}
and formulate the optimization objective as
\[
\max \, \{ L_{adv} = \mathbb{E}[L_{s} - \alpha L_{DS} - \gamma L_{reg}^{env}] \},
\]
through which we obtain the strengthened environment features $h_{e}'$. Afterward, we obtain
\[
\|h_s - h_e'\| \gg \|h_s - h_e\|.
\]
These expressions indicate that, after strengthening the adversarial augmenter, the strengthened environment features become further separated from both the stable features and the original graph in the latent space. This separation enhances feature discriminability, thereby improving the model's overall ability to distinguish invariant representations.

\section{Experimental Results} \label{sec: experiment}

In this section, we present the experimental results of our proposed method. We introduce the datasets and baselines employed in our experiments and provide a detailed analysis of the experimental results, offering insights into the performance and efficacy of our proposed approach. To gain a deeper understanding of the contributions of individual components, we conduct an ablation study. Additionally, we perform a visualization analysis to provide a comparative assessment of MPAIACL and AIA. We also conduct the hyperparameter experiment of MPAIACL. Notice that all baselines adhere to their original settings, and all experiments are conducted on a single NVIDIA GeForce RTX 4060 Ti 16 GB GPU.

\subsection{Settings}

\textbf{Datasets}. We evaluate our model on the Open Graph Benchmark (OGB)  \cite{hu2020open} (including the Molbbbp, Molbace, Molsider, Moltox21, Moltoxcast, and Molclinxtox) and GOOD \cite{gui2022good} datasets (including MolHiv, Motif, and CMNIST). Notice that Molbbbp, Molbace, Molsider, Moltox21, Moltoxcast, and Molclinxtox are molecular datasets collected from MoleculeNet \cite{wu2018moleculenet}. We utilize the above datasets to conduct covariant shift experiments. We also utilize the GOOD datasets to conduct correlation shift experiments. For the molecular datasets, we adopt the covariate shifts introduced in \cite{sui2024unleashing}, which leveraged scaffold and graph size to create diverse types of covariate shifts. Note that the scaffold-based construction employs scaffold splitting to partition the data into training, validation, and test sets. In contrast, the size-based shift involves training on large graphs, while using smaller graphs for validation and testing. For Molhiv, Motif, and CMNIST, we utilize the covariate shift in GOOD \cite{gui2022good}. The details of the covariant datasets are shown in Table \ref{table:statis info of covariant} and Table \ref{table:statis info of covariate of GOOD}.  For the correlation shift experiment, we utilize Molhiv, Motif, and CMNIST as our datasets. We utilize the correlation shift in GOOD \cite{gui2022good}. The details of the correlation datasets are shown in Table \ref{table:statis info of correlation}.

\textbf{Baselines}. We compare the performance of MPAIACL against a range of representative baselines. For (I) generalization, we compare our method to (1) ERM and IRM \cite{arjovsky2019invariant}, with a focus on identifying which properties of the training data correspond to spurious correlations and which properties capture the phenomenon of interest. (2) VREx \cite{krueger2021out}, which proposes a penalty on the variance of training risks as a simpler alternative. (3) GroupDRO \cite{sagawadistributionally}, combining group DRO models with enhanced regularization. We compare the above general methods because they overlook the importance of tackling stable features and environments from a contrastive perspective. For (II) graph generalization algorithms, we compare our approach with (1) CIGA \cite{chen2022learning}, which aims to capture graph invariance for guaranteed out-of-distribution (OOD) generalization under diverse distribution shifts; (2) DIR-GNN \cite{wu2022discovering} develops a framework for discovering invariant rationales, enabling the construction of inherently interpretable graph neural networks; (3) GSAT \cite{miao2022interpretable} addresses the limitations of post-hoc interpretation methods that often fail to provide stable and reliable explanations and instead extract features that are spuriously correlated with the task by attention mechanisms. (4) GALA \cite{chen2024does} is a novel model that learns invariant graph representations under the guidance of an environment assistant model. We compare the above graph generalization methods because they overlook the covariate shift distribution problem in OOD datasets. For (III) graph augmentation algorithms, we compare our approach with (1) FLAG \cite{kong2022robust}, which iteratively augments node features with gradient-based adversarial perturbations during training; (2) GREA \cite{liu2022graph} separates rationales from environments and learns representations of real and augmented examples in latent spaces, enabling effective graph learning. We compare the above data augmentation methods because they often fail to preserve the stable features, as they lack explicit constraints and may inadvertently destroy the very information. (3) DropEdge \cite{rong2019dropedge} randomly removes edges from the graph. (4) GraphCL \cite{you2020graph} employs a node-dropping and feature-masking strategy during the DA stage. Notice that we only use the mask feature strategy in GraphCL. We also compare with (5) AIA \cite{sui2024unleashing}, which employs a data augmentation strategy to mitigate covariate shifts on graphs, since our proposed MPAIACL is a refined model of it.

\textbf{Evaluation metric}. We adopt several evaluation metrics on different datasets. For the molecular datasets, including Molbbbp, Molbace, Molsider, Moltox21, Moltoxcast, and Molclinxtox, we employ the ROC-AUC as the evaluation metric. For the GOOD dataset, which comprises MolHiv, Motif, and CMNIST, we use ROC-AUC to evaluate MolHiv, and accuracy to evaluate both CMNIST and Motif.

\subsection{Main Results}

\textbf{Covariate shift distribution}. In the covariate shift distribution experiment, we employ the data manipulation approach introduced in \cite{sui2024unleashing}, which induces covariate shifts in the molecular datasets by design. In this experiment, we compared our method MPAIACL with $\*$ different baselines in $\*$ different datasets. In Table \ref{table: result of molecular datasets}, we make comparisons with various baselines in the covariate shift distribution using the OGB benchmark. Most baselines in generalization and graph generalization fail in covariate shift, such as G-DRO, GALA, etc. For the graph augmentation method, DropEdge, GraphCL fails in covariate shift compared with ERM. FLAG, AIA, and MPAIACL are obtaining an improvement to an extent. We consider Molbbbp in the size domain as an example, compared with ERM. For generalization, VREx and G-DRO obtain 0.47\% and 0.98\%, improvement, respectively. 
For graph generalization, CIGA, DIR-GNN, GALA perform $\downarrow$ 12.31\%, $\downarrow$ 1.89\%, and $\downarrow$ 9.35\%, respectively. GSAT performs $\downarrow$ 2.66\%. For graph augmentation, GREA, DropEdge, and GraphCL perform $\downarrow$ 0.95\%, $\downarrow$ 2.45\%, and $\downarrow$ 13.14\%, respectively. FLAG, AIA, and MPAIACL, respectively, obtain 0.97\%, 2.74\%, and 4.47\% improvement. In terms of overall results, MPAIACL demonstrates impressive performance on molecular datasets.  Overall, our results demonstrate that MPAIACL is an effective variant of AIA, offering improved performance in the presence of covariate shifts.

In Table \ref{table: result of GOOD datasets}, we make comparisons with various baselines in covariant shift distribution using the GOOD benchmark.
Many baselines fail in GOOD datasets in covariate shifts. We consider motifs in the size domain as an example, compared with ERM. For generalization, IRM performs $\downarrow$ 0.33\%.  VREx, and GroupDRO obtain 0.93\%, and  0.21\% improvement, respectively. For graph generalization, CIGA performs $\downarrow$ 2.60\%
DIRGNN, GSAT, and GALA obtain 0.53\%, and 1.46\%, 3.04\% improvement, respectively. For graph augmentation, FLAG, DropEdge, and GraphCL perform $\downarrow$ 0.08\%, $\downarrow$ 16.87\% and $\downarrow$ 18.53\%. GREA, AIA, and MPAIACL have 2.39\%, 4.11\%, and 11.97\%  improvement, respectively. Overall, MPAIACL delivers strong performance in both Motif and MolHiv, thereby validating its effectiveness. However, in the CMNIST dataset, GALA achieves the best performance, while MPAIACL underperforms compared to GALA. In summary, MPAIACL achieves the best comprehensive performance in the GOOD dataset compared with other baselines.

\textbf{Correlation shift distribution}. While our primary focus is on addressing the covariate shift distribution problem, we also assess the performance of MPAIACL under correlation shift. Following the experimental setup in GOOD \cite{gui2022good}, we create correlation shift scenarios using the MolHiv (size, scaffold) and Motif (size, basis) datasets. For a comprehensive evaluation, MPAIACL was compared against 11 baseline methods: (1) IRM, (2) ERM \cite{arjovsky2019invariant}, (3) VREx \cite{krueger2021out}, (4) G-DRO \cite{sagawadistributionally}, (5) CIGA \cite{chen2022learning}, (6) DIR-GNN \cite{wu2022discovering}, (7) GSAT \cite{miao2022interpretable}, (8) FLAG \cite{kong2022robust}, (9) AIA \cite{sui2024unleashing}, (10) DropEdge \cite{rong2019dropedge}, and (11) GraphCL \cite{you2020graph}. The results are presented in Table \ref{table: result of concept shift distribution}. We can observe that most methods are effective in handling Correlation shift distributions, with the notable exceptions of DropEdge and GraphCL, which employ random graph augmentation techniques. Comparing MPAIACL with the other baselines, we can observe that MPAIACL has superior performance across different domains in MolHiv, Motif, and CMNIST. Overall, the result also demonstrates that MPAIACL is effective in the correlation issue, which demonstrates that MPAIACL is an improved version of AIA.

\begin{table*}[]
\caption{Experimental results of covariate shift distribution of molecular datasets. ``-'' indicates that GALA's sampling strategy did not achieve multilabel binary classification. The best-performing result is highlighted in \textbf{bold}, while the second-best result is indicated with \underline{underlining}.}
\small
\renewcommand\arraystretch{1.5}
\setlength{\tabcolsep}{3pt}
\scalebox{0.7}{
\begin{tabular}{clcccccccccccc}
\hline
\multirow{2}{*}{\textbf{Type}} &
  \multicolumn{1}{c}{\multirow{2}{*}{\textbf{Method}}} &
  \multicolumn{2}{c}{\textbf{Molbbbp}} &
  \multicolumn{2}{c}{\textbf{Molbace}} &
  \multicolumn{2}{c}{\textbf{Molsider}} &
  \multicolumn{2}{c}{\textbf{Moltox21}} &
  \multicolumn{2}{c}{\textbf{Moltoxcast}} &
  \multicolumn{2}{c}{\textbf{Molclintox}} \\ \cline{3-14} 
 &
  \multicolumn{1}{c}{} &
  \textbf{size} &
  \textbf{scaffold} &

  \textbf{size} &
  \textbf{scaffold} &
  \textbf{size} &
  \textbf{scaffold} &
  \textbf{size} &
  \textbf{scaffold} &
  \textbf{size} &
  \textbf{scaffold} &
  \textbf{size} &
  \textbf{scaffold} 
  \\ \hline
General        & IRM     & 77.56$\pm$2.48 & 67.22$\pm$1.15   & 77.06$\pm$1.65 & 69.15$\pm$2.59 & 54.20$\pm$1.26 & 55.10$\pm$1.32 & 71.26$\pm$0.76 & 68.72$\pm$1.71 & 57.65$\pm$3.08 & 57.37$\pm$0.73 & 62.59$\pm$8.91 &  58.68$\pm$3.33 \\
generalization & ERM     & 78.29$\pm$3.76 & 68.10$\pm$1.68   & 82.22$\pm$0.79 & 76.62$\pm$1.20 & 56.82$\pm$1.95 & 55.50$\pm$1.52 & 70.43$\pm$1.53 & 74.07$\pm$0.36 & 62.98$\pm$1.37 & 63.34$\pm$0.51 & 87.59$\pm$5.38 & 85.94$\pm$1.77 \\ 
& VREx     & 78.76$\pm$2.37 & 68.74$\pm$1.03    & 79.67$\pm$0.23 &  66.70$\pm$0.14 & 56.82$\pm$1.95 & 58.79$\pm$0.23 & 63.22$\pm$0.25 & 68.94$\pm$0.76 & 56.79$\pm$1.44& 57.83$\pm$0.92 & 80.59$\pm$9.56 & 79.10$\pm$1.13\\
& G-DRO     & 79.27$\pm$2.43  & 66.47$\pm$2.39   & 79.64$\pm$0.14& 67.10$\pm$0.11 & 54.78$\pm$0.80  & 57.91$\pm$0.25 & 61.82$\pm$1.49 & 69.34$\pm$0.30 &59.36$\pm$1.79 & 58.80$\pm$0.07& 87.00$\pm$4.11 & 79.06$\pm$0.73  \\

\hline
Graph          & CIGA    & 65.98$\pm$3.31 & 64.92$\pm$2.09   & 68.46$\pm$2.17 & 74.39$\pm$2.19 & 52.13$\pm$1.71 & 50.44$\pm$1.17 & 64.86$\pm$2.19 & 58.27$\pm$2.32 & 52.04$\pm$2.15 & 55.29$\pm$2.19 & 75.42$\pm$1.43 & 58.95$\pm$6.77 \\
generalization & DIR-GNN & 76.40$\pm$4.43 & 66.86$\pm$2.25   &77.48$\pm$2.28    & 77.98$\pm$2.81  & 54.02$\pm$0.61 & 52.47$\pm$1.57 & 68.52$\pm$2.13  & 67.91-$\pm$1.21 & 62.75$\pm$0.65 & 56.18$\pm$0.86 & 36.88$\pm$1.24 & 71.51$\pm$5.18 \\
               & GSAT    & 75.63$\pm$3.83 & 66.78$\pm$1.45   & 78.09$\pm$2.19 & 73.84$\pm$3.05 & 56.67$\pm$2.10& \textbf{61.27$\pm$0.42} & 68.75$\pm$4.44 & 73.20$\pm$0.51 & 62.64$\pm$1.01 & 61.80$\pm$0.43 & 87.73$\pm$3.37 & \textbf{89.52$\pm$1.77} \\ 
               &GALA &68.94$\pm$8.41&57.80$\pm$2.35&76.70$\pm$6.44&70.93$\pm$2.88&-&-&-&-&-&-&-&-\\
               \hline
               & FLAG    & 79.26$\pm$2.26 & 67.69$\pm$2.36   &\underline{83.56$\pm$0.80} & 76.93$\pm$ 0.93& 52.09$\pm$1.44 & 58.65$\pm$2.03 & 65.72$\pm$1.12 & \underline{75.88$\pm$0.90} & \underline{65.98$\pm$2.52} & 62.81$\pm$0.18 & 84.99$\pm$5.56 & 86.83$\pm$1.35  \\
Graph          & GREA    & 77.34$\pm$3.52 & 67.65$\pm$1.89 & 83.18$\pm$0.46 & \underline{78.78$\pm$1.55} & \underline{57.40$\pm$0.64} & \underline{59.53$\pm$1.05} & \textbf{75.68$\pm$0.29} & \textbf{76.11$\pm$0.73} & 65.28$\pm$0.62 & \textbf{65.83$\pm$0.60} & \underline{90.40$\pm$4.02}& 88.37$\pm$2.66 \\
augmentation   & AIA     & \underline{81.03$\pm$5.15} & \underline{68.03±1.52}   & 80.76$\pm$1.19 & 77.74$\pm$2.36 & 54.41$\pm$2.53 & 56.32$\pm$0.32 & 70.29$\pm$1.04 & 75.37$\pm$0.68 & 65.13$\pm$0.45 & \underline{63.72$\pm$0.70} & 82.49$\pm$15.83 & 87.87$\pm$1.81  \\
& D-Edge     &75.84$\pm$0.31  &  60.03$\pm$0.70    & 74.39$\pm$0.20 & 70.12$\pm$0.20 &55.12$\pm$0.25 & 48.09$\pm$0.09 & 60.98$\pm$1.52 & 68.72$\pm$0.29 & 61.07$\pm$1.20& 57.19$\pm$0.90& 82.98$\pm$2.70&  78.13$\pm$5.36  \\
& GraphCL     &65.15$\pm$0.87  & 61.13$\pm$2.26   & 69.29$\pm$3.43 &62.95$\pm$1.37 & 49.48$\pm$0.08& 47.24$\pm$0.10& 49.09$\pm$0.15& 47.07$\pm$0.54 & 49.26$\pm$0.06& 50.06$\pm$0.10& 43.36$\pm$1.71& 47.56$\pm$2.31 \\

\rowcolor{light-gray}
 &
   MPAIACL & \textbf{82.76$\pm$0.03}
   &\textbf{69.64$\pm$0.70}
   &\textbf{83.68$\pm$0.04}

   &\textbf{79.11±0.54}
  &\textbf{57.54$\pm$0.02}&57.86±1.02&\underline{71.53±0.62}&74.36±0.31&\textbf{66.25$\pm$0.34}&63.34$\pm$0.36&\textbf{93.75$\pm$0.01}& \underline{87.90±1.40} \\ \hline

\end{tabular}
}

\label{table: result of molecular datasets}
\end{table*}

\begin{table*}[]
\caption{Experiment results of covariant shift distribution of GOOD datasets. The best-performing result is highlighted in \textbf{bold}, while the second-best result is indicated with \underline{underlining}.}
\footnotesize
\renewcommand\arraystretch{1.5}
\setlength{\tabcolsep}{6pt}
\begin{tabular}{clccccccc}
\hline
\multirow{2}{*}{\textbf{Type}} &
  \multicolumn{1}{c}{\multirow{2}{*}{\textbf{Method}}} &
  \multicolumn{2}{c}{\textbf{Motif}} &
  \multicolumn{2}{c}{\textbf{MolHiv}} &
  \multicolumn{1}{c}{\textbf{CMNIST}} &
 \\ \cline{3-8} 
 &
  \multicolumn{1}{c}{} &
  \textbf{size} &
  \textbf{basis} &
  \textbf{size} &
  \textbf{scaffold} &
  \textbf{color} &

  \\ \hline
General        & IRM     &51.41$\pm$3.78&61.52$\pm$7.11 &59.00$\pm$2.92&67.97$\pm$1.84&27.83$\pm$2.13 \\
generalization & ERM    &51.74$\pm$2.88&68.66$\pm$4.25&\underline{59.94$\pm$2.37}&69.58$\pm$2.51& 28.60$\pm$1.87 \\ 
& VREx     &52.67$\pm$5.54&40.49$\pm$5.66&58.53$\pm$2.88&66.62$\pm$2.55&28.48$\pm$2.87\\
& G-DRO     &51.95$\pm$5.86&68.24$\pm$8.92&58.98$\pm$2.16&69.17$\pm$0.85& 29.07$\pm$3.14\\

\hline
Graph          & CIGA    &49.14$\pm$8.34&66.43$\pm$11.31&59.55$\pm$2.56&69.40$\pm$2.39&32.22$\pm$2.67 \\
generalization & DIR-GNN &52.27$\pm$4.56&62.07$\pm$8.75&58.08$\pm$2.31&68.07$\pm$2.29&33.20$\pm$6.17 \\
               & GSAT    &53.20$\pm$8.35&62.80$\pm$11.41&58.06$\pm$1.98&68.66$\pm$1.35&28.17$\pm$1.26 \\
               &GALA &54.78$\pm$6.05&55.03$\pm$10.00&58.76$\pm$1.86&67.37$\pm$4.58& \textbf{53.30$\pm$2.32}&\\\hline
               & FLAG    &51.66$\pm$4.14&61.12$\pm$5.39&59.54$\pm$1.27&68.45$\pm$2.30&32.30$\pm$2.69  \\
Graph          & GREA    &\underline{54.13$\pm$10.02}&56.74$\pm$9.23& 52.77$\pm$0.18&67.79$\pm$2.56&29.02$\pm$3.26 \\
augmentation  
& AIA     &55.85$\pm$7.98&\underline{68.93$\pm$0.21}&56.30±1.42&\textbf{71.15$\pm$1.81}& 36.37$\pm$4.44  \\
& D-Edge     &34.87$\pm$1.08&33.90$\pm$0.10&55.76$\pm$3.02&56.32$\pm$0.97&15.07$\pm$1.99  \\
& GraphCL     &33.21$\pm$2.51&33.11$\pm$7.85&48.87$\pm$2.27&50.71$\pm$3.71&10.39$\pm$1.10\\
\rowcolor{light-gray}
 &
   MPAIACL 
  &\textbf{63.71$\pm$5.02}&\textbf{69.77$\pm$0.79}&\textbf{60.39$\pm$4.01}&\underline{69.69±2.11}& \underline{42.35$\pm$0.14}& \\ \hline

\end{tabular}

\label{table: result of GOOD datasets}
\end{table*}

\begin{table*}[]
\caption{Experiment results of correlation shift distribution. The best-performing result is highlighted in \textbf{bold}, while the second-best result is indicated with \underline{underlining}.}
\footnotesize
\renewcommand\arraystretch{1.5}
\begin{tabular}{clccccccc}
\hline
\multirow{2}{*}{\textbf{Type}} &
  \multicolumn{1}{c}{\multirow{2}{*}{\textbf{Method}}} &
  \multicolumn{2}{c}{\textbf{Motif}} &
  \multicolumn{2}{c}{\textbf{MolHiv}} &
 \\ \cline{3-7} 
 &
  \multicolumn{1}{c}{} &
  \textbf{size} &
  \textbf{basis} &
  \textbf{size} &
  \textbf{scaffold} &

  \\ \hline
General        & IRM     &64.68$\pm$0.75& 74.79$\pm$1.01&53.97$\pm$1.77&69.07$\pm$0.35\\
generalization & ERM    &65.59$\pm$0.14&81.37$\pm$0.33&52.24$\pm$1.73& 70.27$\pm$1.49\\ 
& VREx     &51.90$\pm$3.85&56.18$\pm$2.32&71.17$\pm$8.39&65.68$\pm$1.76\\
& G-DRO    &64.69$\pm$0.98& 75.30$\pm$0.23&50.81$\pm$1.45&70.28$\pm$1.02\\

\hline
Graph          & CIGA    &51.95$\pm$2.45&68.87$\pm$7.22& 73.20$\pm$0.34& \textbf{71.95$\pm$0.87} \\
generalization & DIR-GNN & 47.12$\pm$7.42&75.07$\pm$6.66&69.72$\pm$2.28&67.40$\pm$2.23\\
               & GSAT    & 37.33$\pm$1.01&52.20$\pm$0.20&49.08$\pm$5.07&67.56$\pm$0.82\\ 
              
               \hline
               & FLAG   &\underline{66.78$\pm$0.70}&\underline{79.04$\pm$0.43}&\textbf{73.17$\pm$1.84}&69.37$\pm$0.84 \\
Graph         

augmentation   & AIA     &65.42$\pm$2.54&76.26$\pm$5.02&70.37$\pm$0.43&69.81$\pm$1.18 \\
& D-Edge     &37.31$\pm$1.79&34.90$\pm$3.40&59.60$\pm$1.25&66.60$\pm$1.13 \\
& GraphCL     &36.05$\pm$1.50&37.68$\pm$1.57&57.03$\pm$2.57&58.88$\pm$5.41\\
\rowcolor{light-gray}
 &
   MPAIACL & \textbf{69.44$\pm$1.71}
  &\textbf{81.64$\pm$1.65}&70.40$\pm$1.33&  \underline{71.78$\pm$0.87} \\ \hline

\end{tabular}

\label{table: result of concept shift distribution}
\end{table*}

\begin{figure}[]
    \centering
    \vspace{0.6cm} 
    \includegraphics[scale=0.4]{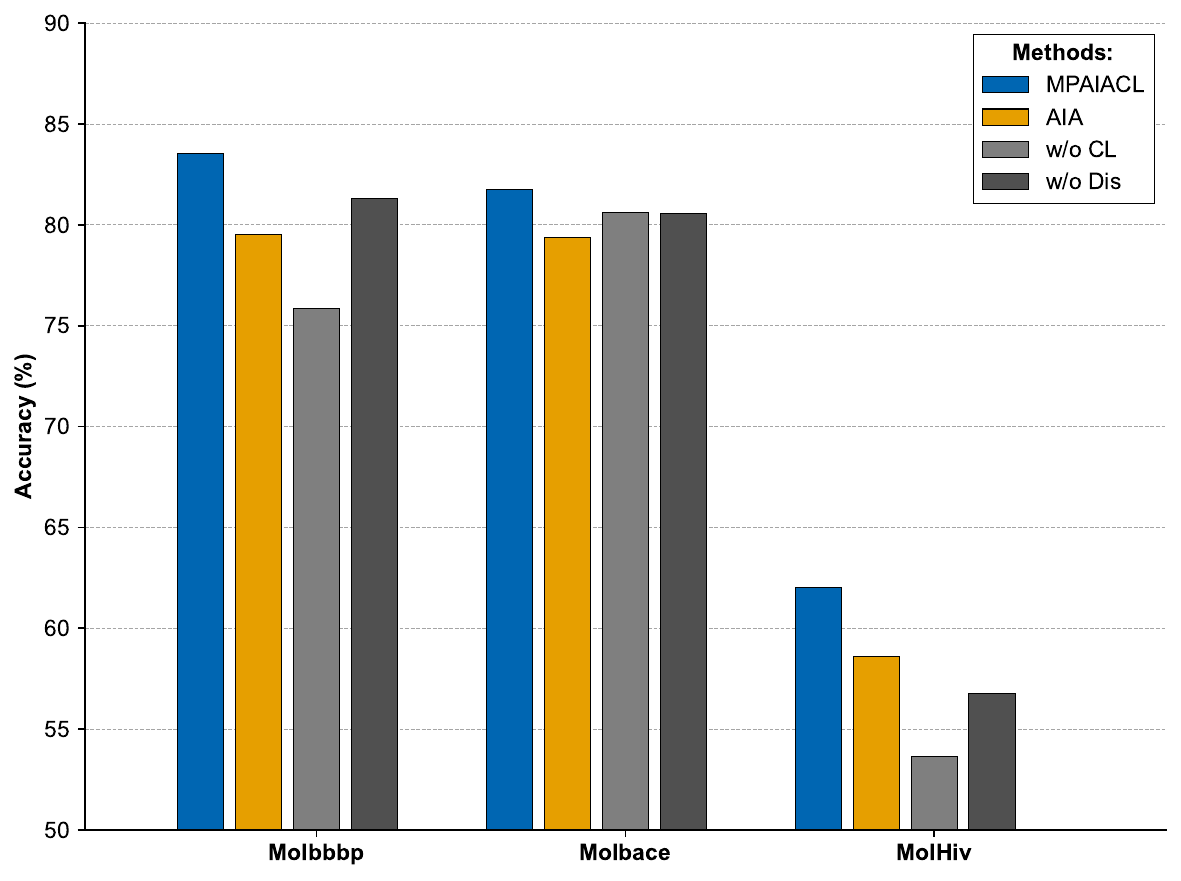}
    \caption{Ablation study on three different datasets. This demonstrates that the performance of MPAIACL outperforms others.}
    \label{fig: ablation study}
\end{figure} 

\subsection{Ablation Study}

We conduct an ablation study to investigate the contributions of two key components in our approach: contrastive learning in the stable features generator and the use of Wasserstein distance in the adversarial augmentation. 
As illustrated in Figure \ref{fig: ablation study}, we denote the following variants of MPAIACL: (1) ``w/o cl'': MPAIACL without contrastive learning in the stable features generator; (2) ``w/o dis'': MPAIACL without the Wasserstein distance-based regularization between stable features, original graph, and environment features in the adversarial augmenter. We observe that the performance of each component degrades significantly when used independently, often resulting in worse performance than even AIA. Taking Molbbbp as an example, ``w/o cl'' is even worse than AIA, which performs $\downarrow$ 3.68\%. Although ``w/o dis''  1.78\% improvement, it perform $\downarrow$  2.23\% than MPAIACL, respectively. 
In contrast, the combined version of MPAIACL achieves superior performance, obtaining 4.01\% improvement over AIA, highlighting the importance of integrating these components for effective covariate shift adaptation. Without incorporating vector information in the latent space, the vectors would approach each other without limitation, leading to suboptimal performance. This highlights the importance of utilizing vector information to regulate the latent space.  Without utilizing the Wasserstein distance between stable features and environment features, the performance of our approach still falls short of the combined version. To achieve optimal fine-tuning performance, it is also essential to strengthen the adversarial augmenter.

\begin{figure*}[]
    \centering
    \subfigure[Line chart of temperature $\tau$]{
        \label{fig: temp}
        \includegraphics[scale = 0.25]{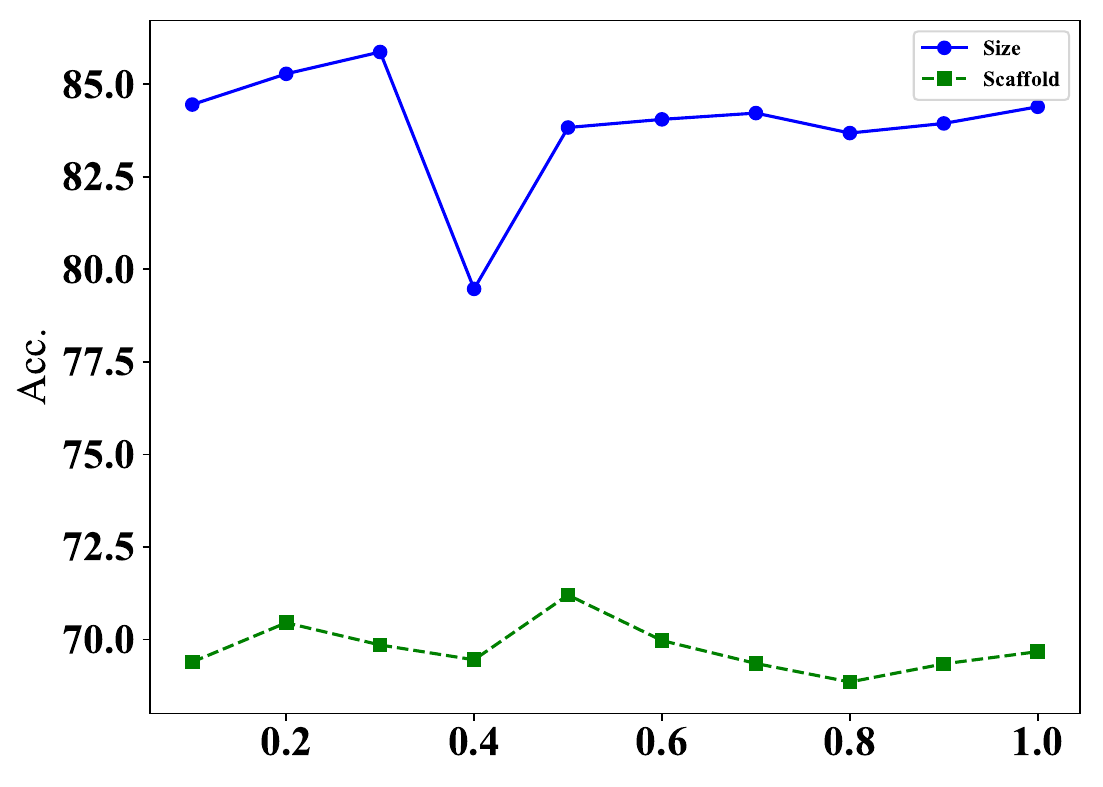}
    }
    \subfigure[Line chart of $\lambda$]{
        \label{fig: label}
        \includegraphics[scale = 0.25]{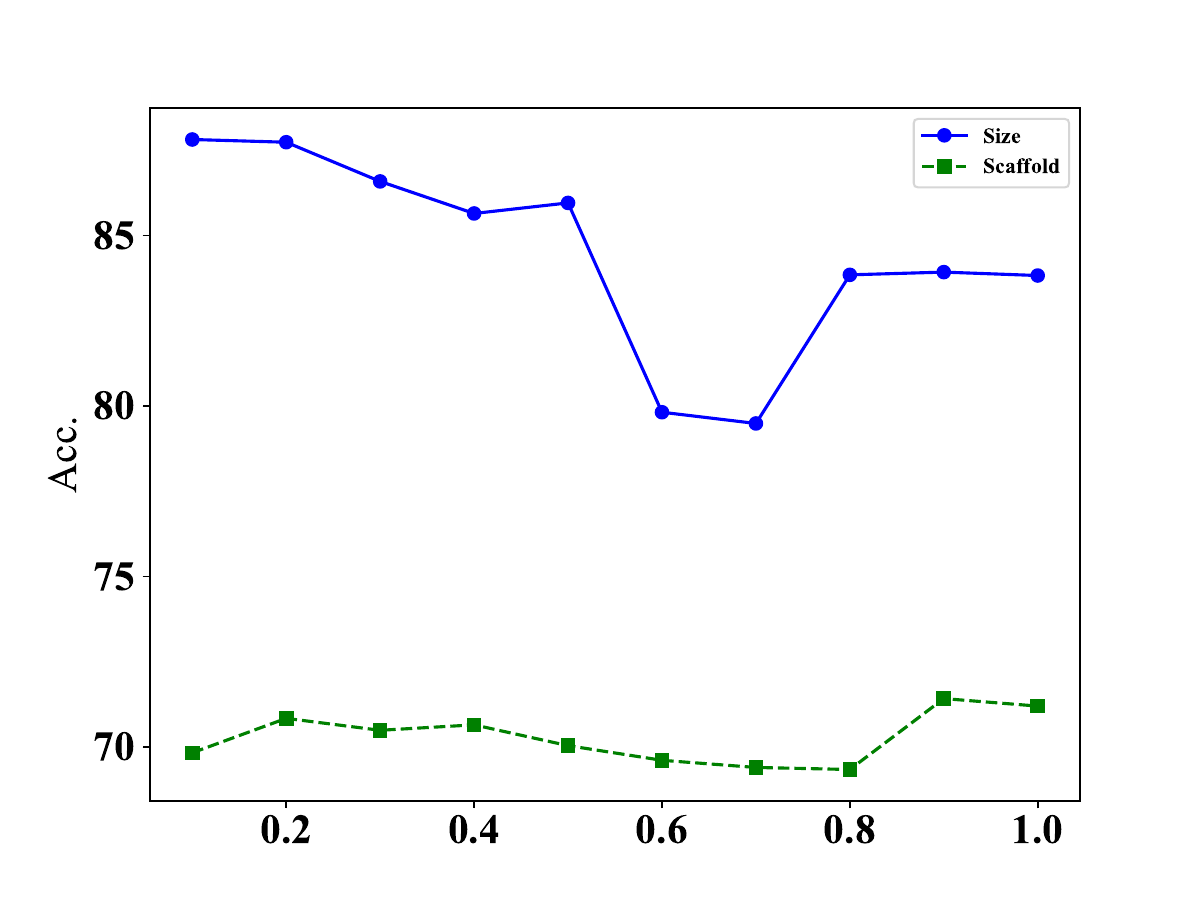}
    }
    \subfigure[Line chart of $\alpha$]{
        \label{fig: dis}
        \includegraphics[scale = 0.25]{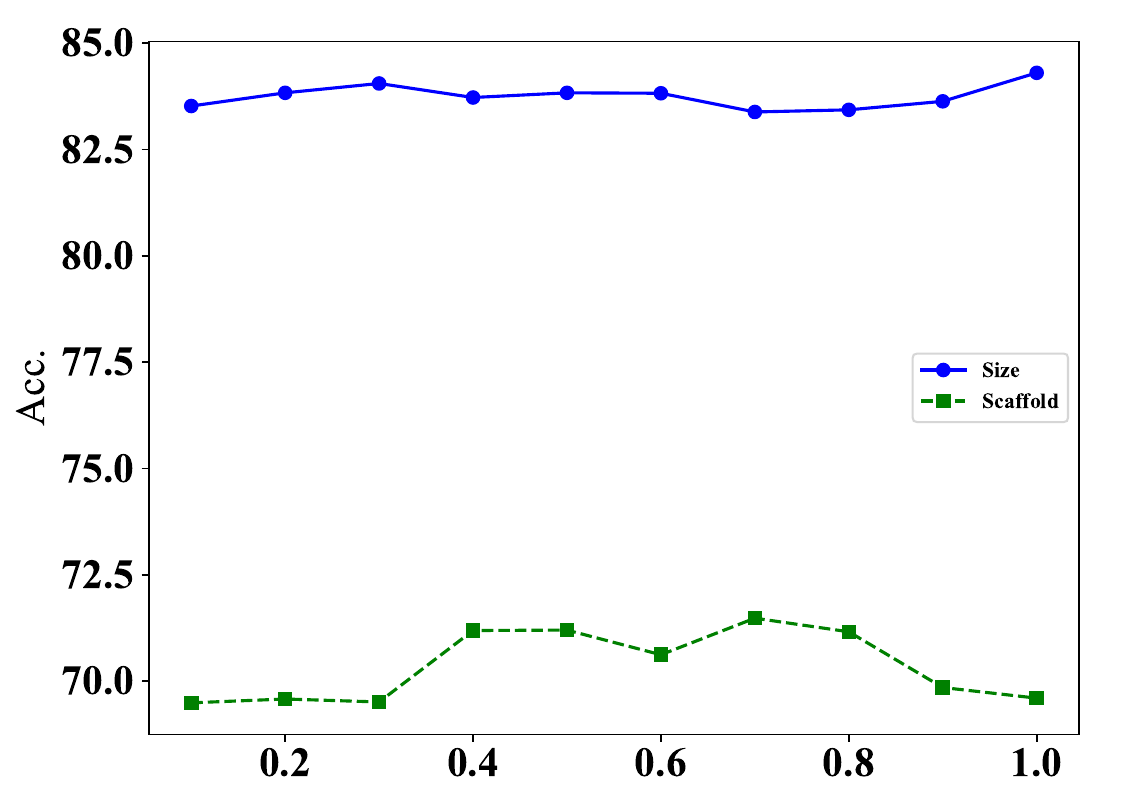}
    }  
    \caption{Hyperparameter analysis.}
    \label{fig: hyperparameter}
\end{figure*}

\begin{figure}[]
    \flushleft 
    \subfigure[Original]{
        \label{fig: original_vis_1}
        \includegraphics[scale = 0.15]{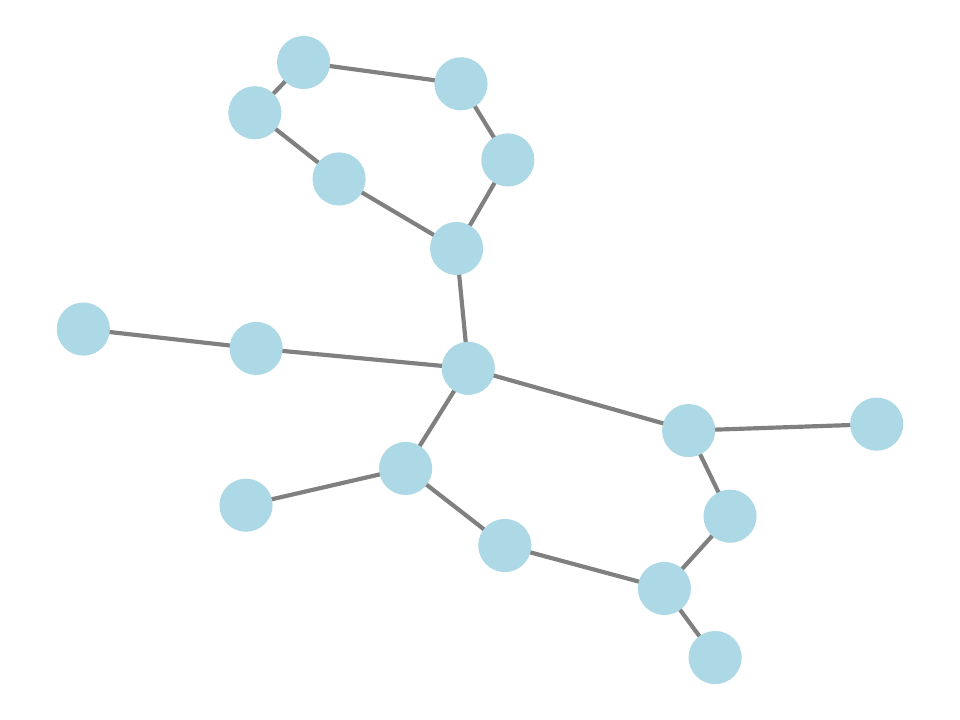}
    }
    \subfigure[AIA]{
        \label{fig: AIA_vis_1}
        \includegraphics[scale = 0.15]{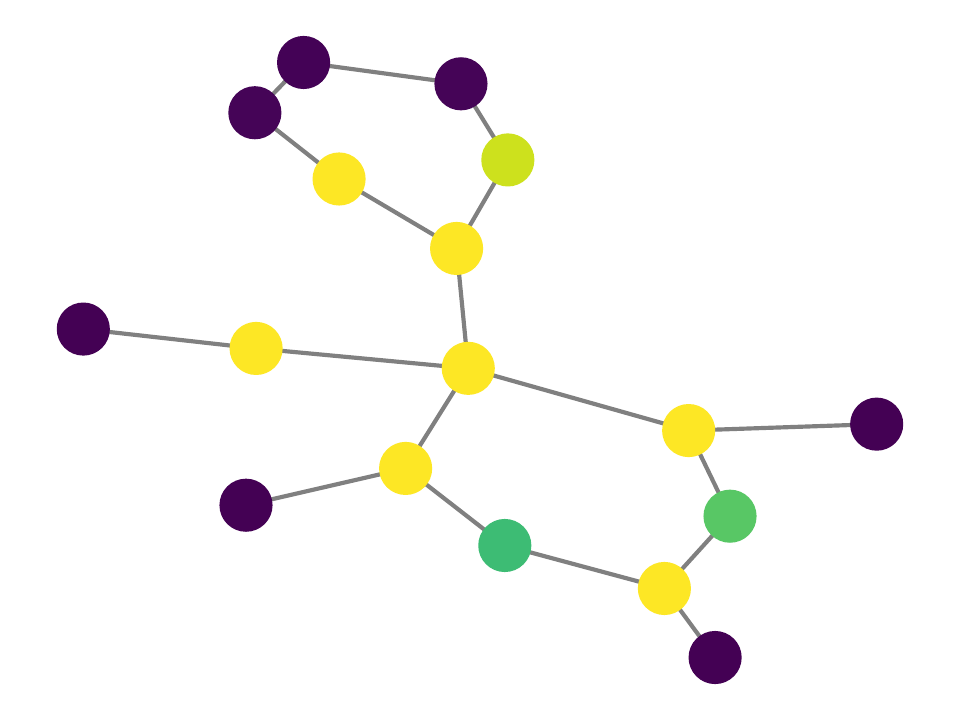}
    }
    \subfigure[MPAIACL]{
        \label{fig: MPAIACL_vis_1}
        \includegraphics[scale = 0.15]{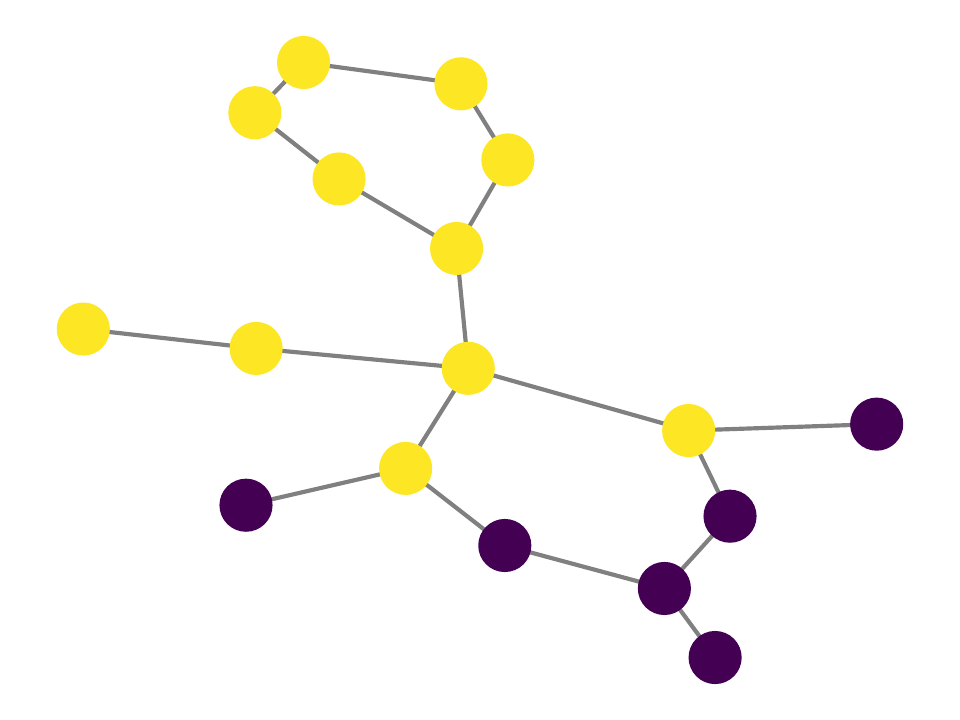}
    }

    \subfigure[Original]{
        \label{fig: original_vis_2}
        \includegraphics[scale = 0.15]{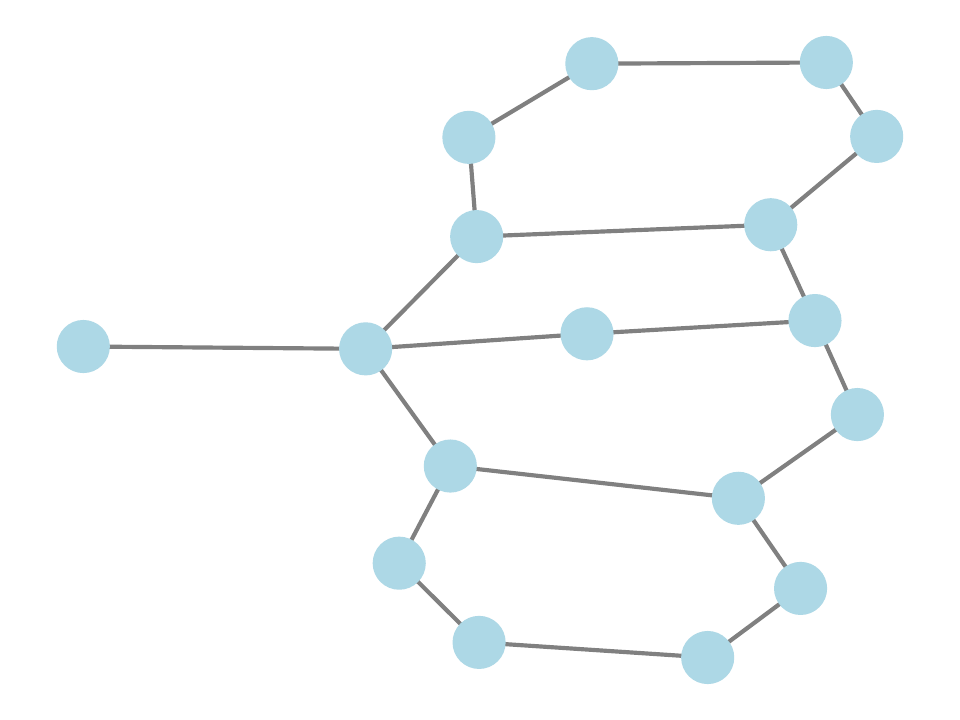}
    }
    \subfigure[AIA]{
        \label{fig: AIA_vis_2}
        \includegraphics[scale = 0.15]{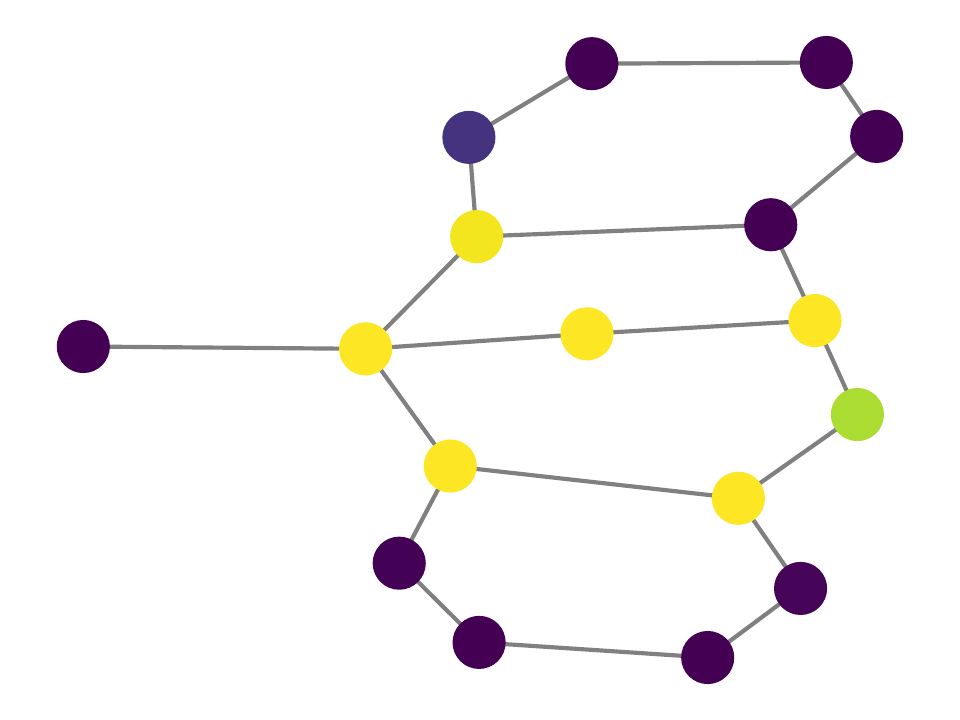}
    }
    \subfigure[MPAIACL]{
        \label{fig: MPAIACL_vis_2}
        \includegraphics[scale = 0.15]{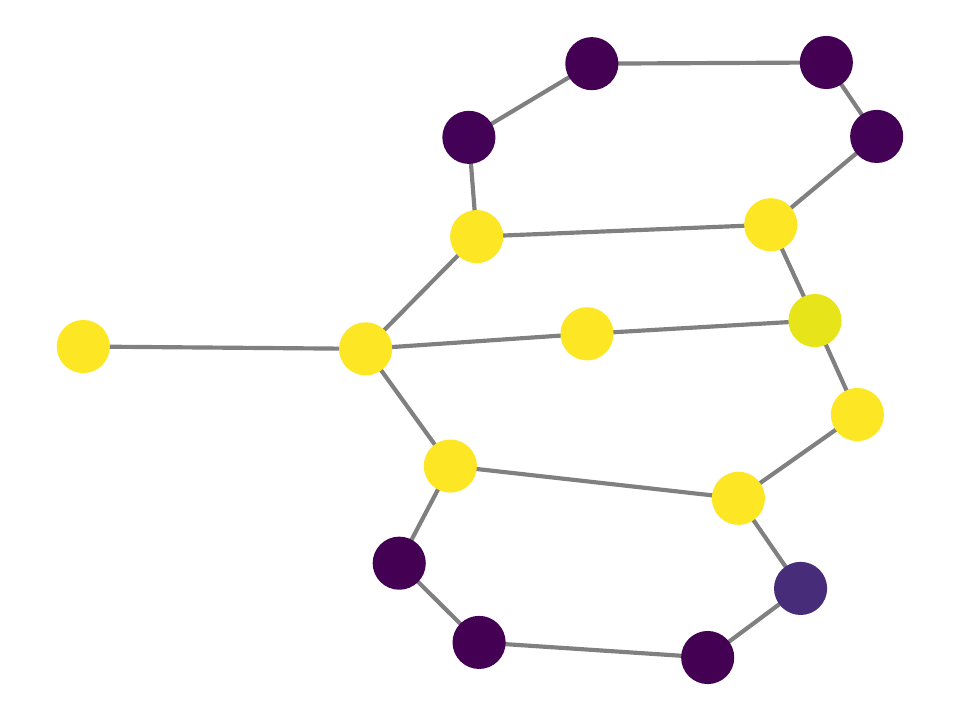}
    }
    
    \caption{Visualization.}
    \label{fig: visuilization}
    
\end{figure}

\subsection{Visualization}

Here, we provide a visualization of MPAIACL and AIA to intuitively demonstrate the ability of MPAIACL to capture stable features. We utilize \textit{NetworkX} to generate visualizations of our results. As shown in Figure \ref{fig: visuilization}, the visualization is organized as follows: the left column displays the original graphs selected from the Molbbbp training set, the middle column shows the stable features captured by AIA, and the right column presents the stable features captured by MPAIACL. In these figures, the color intensity represents the stable degree; the brighter the color, the greater the stability of the features. We can see that MPAIACL is generally a match for AIA, which indicates that MPAIACL is an effective method for capturing stable features. Furthermore, MPAIACL appears to exhibit a more optimistic outlook on stable features and tends to capture more structure as stable features. While MPAIACL outperforms AIA in terms of accuracy, a crucial consideration is whether it also captures more consistent and stable features. As a variant of AIA, MPAIACL leverages latent information to enhance performance, but its approach is more empirical, lacking a solid theoretical foundation. We will investigate this question further in our future work.

\subsection{Hyperparameter Analysis}

In this subsection, we conduct experiments to investigate hyperparameters' impact on our method's performance. Specifically, we analyze the influence of the temperature parameter $\tau$ in Equation \ref{equation: contrastive loss}, the regularization strength $\lambda$ in Equation \ref{equation: obj of stable generator}, and the hyperparameter $\alpha$ in Equation \ref{equation: obj of Adversarial Augmenter}. We use Molbbbp as our dataset with different domains (size, scaffold) and explore the performance when the hyperparameter varies from 0 to 1 in increments of 0.1. It is noticed that the default hyperparameter settings are $\tau$ = 0.5, $\lambda$ = 1, and $\alpha $ = 0.5. 

As illustrated in Figure \ref{fig: hyperparameter}, we have three different line charts for $\tau$, $\lambda$, and $\alpha$, respectively, in different domains. The performance varies significantly with different values of the hyperparameters. Details are discussed below.

(i) The impact of varying $\tau$ on domain size is illustrated in Figure \ref{fig: temp}. The accuracy increases steadily as $\tau$ ranges from 0 to 0.3. However, a sudden drop in accuracy occurs at $\tau$ = 0.4. Beyond this point, the accuracy remains relatively stable, with minimal fluctuations between $\tau$ equals 0.5 and 1. In the domain scaffold, the accuracy exhibits slight fluctuations throughout the entire range of $\tau$ values. 

(ii) TAs shown in Figure \ref{fig: label}, the impact of varying $\lambda$ on domain size reveals a general downward trend in overall accuracy as $\lambda$ increases from 0.1 to 1. Notably, the accuracy experiences a sharp decline at $\lambda$ = 0.6 but recovers slightly at $\lambda$ = 0.8. In the domain scaffold, the accuracy remains relatively stable, with only minor fluctuations observed throughout the entire range of $\lambda$ values. 

(iii) TFor the change of $\alpha$ in domain both size and scaffold, the accuracy slightly fluctuates throughout the entire range. Across both domain size and scaffold illustrated in Figure \ref{fig: dis}, the accuracy remains relatively stable with respect to the changes of $\alpha$, exhibiting only minor fluctuations throughout the entire range.

\begin{table*}[t]
\caption{Complexity comparison between AIA and MPAIACL.}
\footnotesize
\renewcommand\arraystretch{1.5}
\setlength{\tabcolsep}{4pt}
\begin{tabular}{llllll}
\hline
\multirow{2}{*}{\textbf{Dataset}} &
  \multicolumn{1}{c}{\multirow{2}{*}{\textbf{Domain}}} &
  \multicolumn{2}{c}{\textbf{AIA}} &
  \multicolumn{2}{c}{\textbf{MPAIACL (ours)}} \\ \cline{3-6} 
 &
  \multicolumn{1}{c}{} &
  \multicolumn{1}{c}{\makecell{\textbf{Time} \\ \textbf{consumption}}} &
  \multicolumn{1}{c}{ \makecell{\textbf{Memory} \\ \textbf{consumption}}} &
  \multicolumn{1}{c}{\makecell{\textbf{Time} \\ \textbf{consumption}}} &
  \multicolumn{1}{c}{\makecell{\textbf{Memory} \\ \textbf{consumption}}} \\ \hline
\multirow{2}{*}{\textbf{MolHiv}}  & \textbf{size}     
& 00h 41m 50s & 395.50 MB 
& 0 h 47 m 29 s & 456.41 MB\\
& \textbf{scaffold} 
& 00h 39m 12s & 365.86 MB 
& 0 h 44 m 20 s & 430.39 MB\\ \hline

\multirow{2}{*}{\textbf{Motif}}  & \textbf{size}     
& 00h 14m 32s & 223.65 MB 
& 0 h 16 m 10 s & 268.88 MB  \\
& \textbf{basis}    
& 00h 14m 08s & 225.23 MB  
& 0 h 15 m 29 s & 268.29 MB  \\ \hline

\multirow{2}{*}{\textbf{Molbbbp}} & \textbf{size}     
& 00h 02m 46s & 355.89 MB  
& 0 h 2 m 48 s & 409.58 MB \\
& \textbf{scaffold} 
& 00h 02m 12s & 290.09 MB  
& 0 h 2 m 24 s & 338.61 MB  \\ \hline

\multirow{2}{*}{\textbf{Molbace}} & \textbf{size}     
& 00h 02m 04s & 414.96 MB  
& 0 h 2 m 15 s & 492.09 MB \\
& \textbf{scaffold} 
& 00h 02m 01s & 399.21 MB 
& 0 h 2 m 11 s & 462.71 MB \\ \hline

\textbf{CMMNIST} & \textbf{color}    
& 02h 23m 08s & 2201.90 MB 
& 03h 17m 14s & 2704.77 MB \\ \hline

\end{tabular}

\label{table: complexity comparison}
\end{table*}

\subsection{Complexity Analysis}

We further provide an analysis of the time complexity of our proposed method to assess its efficiency. We also present a comparison of the time and max GPU memory consumption of various datasets used in our experiments in Table \ref{table: complexity comparison}, providing an overview of their computational requirements. As shown, the time and memory consumption of our proposed method are higher than those of AIA. This is because our method, MPAIACL, incurs additional computational costs due to the calculation of contrastive loss in the stable feature generator and the computation of Wasserstein distance in the adversarial augmenter. MPAIACL introduces slightly higher computational overhead compared with AIA in both time and memory consumption. Specifically, the training time increases by approximately 5–15 \% on molecular and motif datasets, while a larger increase is observed on the CMMNIST dataset due to its higher data complexity. Similarly, the memory consumption grows by roughly 15–25 \% across all datasets.

To calculate the time complexity of our proposed method based on the architecture mentioned in Section \ref{sec: Model Architechure}, we first define $n$ and $e$ as the total number of nodes and edges, respectively. Let $B$ denote the batch size. Let $l$, $l_s$, and $l_a$ denote the layers of the GNN backbone, stable feature generator, and adversarial augmenter, respectively. Let $h$, $h_s$, and $h_a$ denote the hidden layers of the GNN backbone, stable feature generator, and adversarial augmenter, respectively. Utilizing the definition above, the time complexity of the stable feature generator is $O(B(2leh$ + $l_seh_s))$. $2leh$ denotes the hidden embedding through two GNN layers with message-passing. $l_seh_s$ denotes generating masks for nodes and edges of graphs. The time complexity regulation term is $O(2Bn)$, since we utilize the result of the stable generator to calculate the label information. The time complexity of the adversarial augmenter is $O(B(2leh$ + $l_aeh_a))$, the same as the stable generator. Its corresponding regularization term is $O(B(n + e))$, since we calculate the Wasserstein distance of $h_{ori}$ with $h_{adv}$. For convenience, we assume that $l_s$ = $l_a$, $h_s$ = $h_a$. The total time complexity of MPAIACL is then $O(2B(2leh + l_s e h_s + 1.5n + e))$.

\subsection{Limitation} \label{sec: limitation}

Although MPAIACL outperforms various baselines in our experiments, we acknowledge and reflect on the limitations of our approach. Although our approach is grounded in the manifold assumption \cite{van2020survey} and intuition behind contrastive learning, we acknowledge that our work relies more heavily on empirical evidence and experimental experience. The strengthened strategy introduces a degree of randomness into the model's performance, particularly in the adversarial augmenter, since we lack knowledge of the ground truth environment embedding and only make the environment embedding far away from the stable one and the augmentation one.  Another limitation is that the augmented environment features may not accurately reflect real-world situations, particularly after contrastive learning, which can exacerbate this discrepancy. We will explore more theoretical ways to make the process more stable.

\section{Conclusion} \label{sec: conclusion}

In this paper, we investigate graph classification under covariate distribution shift. We discover that the representation information in the latent space remains under-explored. Consequently, we harness the representative information in the latent space, thereby unlocking its potential to improve the performance of graph classification. We propose MPAIACL, a method that employs contrastive learning to fine-tune the existing model by fully releasing the power of the information in the latent space. Experiment results demonstrate the effectiveness of MPAIACL as it performs well compared to other baselines in different OOD datasets. Moreover, our in-depth study reveals that leveraging the latent space information is effective, providing valuable insight into the potential benefits of this approach. In future work, we plan to develop a theoretically grounded framework for augmenting environmental features in a principled manner, with the goal of improving model generalization in diverse and dynamic settings. Additionally, we intend to investigate self-supervised learning techniques to effectively mitigate the challenges posed by covariate distribution shifts, particularly in scenarios where labeled data is scarce or unavailable.

\bibliographystyle{plainnat}

\bibliography{main.bib}

\end{document}